\documentclass[final,12pt]{elsarticlemioarxiv}

\usepackage{amssymb}
\usepackage{subfigure}
\usepackage{epstopdf}
\usepackage[x11names,table]{xcolor}
\usepackage{multirow}
\usepackage{hhline}
\usepackage{rotating}
\usepackage{tikz}
\usepackage[margin=1in,footskip=0.7in]{geometry}

\usetikzlibrary{arrows}

\begin{document}

\begin{frontmatter}

\title{Application of Multifractal Analysis to Segmentation of Water Bodies in Optical and Synthetic Aperture Radar Satellite Images}

\author[$1$,$2$,$3$]{Victor Manuel San Martin\corref{cor1}}
\author[$1$,$2$]{Alejandra Figliola}

\address[$1$]{National Scientific and Technical Research Council (CONICET), Argentina.}
\address[$2$]{Institute of Human Development, National University of General Sarmiento, Juan Mar\'{\i}a Guti\'{e}rrez 1150, (1613) Los Polvorines, Buenos Aires, Argentina.}
\address[$3$]{Faculty of Natural Sciences, National University of La Pampa, Uruguay 151, (6300) Santa Rosa, La Pampa, Argentina.}
\cortext[cor1]{Corresponding author. Email address: victor\_sanmartin1@hotmail.com. }

\begin{abstract}
A method for segmenting water bodies in optical and synthetic aperture radar (SAR) satellite images is proposed. It makes use of the textural features of the different regions in the image for segmentation.
The method consists in a multiscale analysis of the images, which allows us to study the images regularity both, locally and globally. As results of the analysis, coarse multifractal spectra of studied images and a group of images that associates each position (pixel) with its corresponding value of local regularity (or singularity) spectrum are obtained.
Thresholds are then applied to the multifractal spectra of the images for the classification. These thresholds are selected after studying the characteristics of the spectra under the assumption that water bodies have larger local regularity than other soil types.
Classifications obtained by the multifractal method are compared quantitatively with those obtained by neural networks trained to classify the pixels of the images in covered against uncovered by water. In optical images, the classifications are also compared with those derived using the so-called Normalized Differential Water Index (NDWI).

\end{abstract}


\end{frontmatter}

\section{Introduction}

In recent decades, remote sensing has become an extremely useful tool for studying the complex dynamics of Earth. This technique allows us to have a huge amount of information of Earth's surface, periodically and with a growing variety.

Along with the rapid development of remote sensing, interest in the study of the geometrical and self-similarity properties of different elements in Earth's topography has grown. Some examples are the study of coastlines, mountain ranges, river basins \cite{don14}, urban sprawls \cite{fra08}, etc. Researchers also have turned their attention to the characteristics of changes introduced in the ground by various environmental disasters such as wildfires \cite{mal05,mor11}, oil spills\cite{tar14}, and others. These studies, which show the intricate nature of the structures present in satellite images captured by different sensors, are based in fractal geometry theory proposed by Mandelbrot \cite{man83} among other authors.

Fractal dimension may be useful when describing binary structures. However, in most cases we find phenomena that cannot be fully described by two states, but adopt values in a given range. To analyze such situations, it is useful to define measures, that is, functions which assign positive real values to subsets of a given set and that are additive \cite{fal04}.

If we consider the subset of the points of a set for which the local measure obeys a power law with respect to its scale with exponent $\alpha$, those subsets could have different fractal dimensions for different $\alpha$ values. We are then in the presence of a multifractal measure. It can be described by its multifractal spectrum, basically the function that associates to each value of $\alpha$ the fractal dimension of the set of points with that exponent. The analysis of structures with such wealth gives us more information than the simple study of a possible unique fractal dimension. In fact, while a study of the exponents with which the measure scales gives us a description of the local regularity of the measure, the multifractal spectrum gives us a global description of the measure.

This dual approach applied to image analysis may be sensitive to the roughness that the analyzed region has locally, while global vision would perceive if this local feature corresponds to an isolated element of the image or it is part of a texture.

Multifractal formalism has been used in the analysis of various complex systems, particularly in image classification, for example \cite{pos09,mil16,vas12,ebr16}. It seems reasonable, then, to analyze remote sensing images using this theory in order to develop tools for automatic segmentation that could allow us to differentiate various ground covers based on their textural features. Particularly, we direct this work towards segmentation of water bodies, because they are of great relevance to meteorology, hydrological studies, quantification and response to natural disasters, etc.

On the other hand, the use of neural networks for image segmentation and classification has become popular. This method, although in most cases computationally slow and dependent on the regions with which you train the network, is highly effective in the classification. It is interesting to use its results to assess the efficiency of segmentation through multifractal analysis.

The present work is organized as follows: Mathematical tools used in the analysis of the images are presented in section \ref{herramientas}, where central concepts of multifractal theory are defined. Besides, the ideas of artificial neuron and neural networks are presented. Section \ref{metodologia} explains how the theoretical concepts of the previous section are adapted for image segmentation and introduces the Normalized Differential Water Index (NDWI) as another method to be applied to optical images. The optical and synthetic aperture radar (SAR) images which will be used as examples of application are described. In section \ref{resultados} the results of the different methods are summarized. Segmentations carried out with the multifractal method are compared with those obtained with neural networks and NDWI for optical images. The conclusions of this research are presented in section \ref{conclusiones}.

\section{Mathematical tools}\label{herramientas}

In this section we define the mathematical tools that we are going to use. This is done also to unify notation. For a deeper explanation of the concepts of measure, fine and coarse multifractal theory, we refer the reader to\cite{fal04,abr09}, while more information on neural networks is contained in \cite{dre05}.

\subsection{Measures on $\mathbb{R}^n$}\label{medidas}

A measure over a set $X\subset\mathbb{R}^n$ is defined as a function which associates a value in $[0,\infty)$ to every subset of $\chi$ \cite{fal04}. The support of a measure ($spt(\mu)$) is defined as the largest closed subset of the set $\chi$ for which every open neighbourhood of every point has positive measure. A finite measure $\mu$ can be considered as a mass distribution over the measure's support.
By definition  a measure is additive, that is: the measure of a union of 2 disjoint subsets is the sum of the measure of the 2 subsets.

The class of Borel sets is defined as the smallest collection of subsets of $X\subset\mathbb{R}^n$ which satisfy:

\begin{enumerate}
\item every open set and every closed set is a Borel set;
\item the union of every finite or countable collection of Borel sets is a Borel set, and the intersection of every finite or countable collection of Borel sets is a Borel set.
\end{enumerate}

If the Borel sets of a set $X$ can be measured with a measure $\mu$, then that measure is called a Borel measure.

In this article we are going to use 'sum measure': let $g(k,l)$ be the intensity in a grey scale of a point $(k,l)$,  and let $\Omega$ be the set of all points in a neighbourhood of width $i$ centered in a point $(m,n)$, then the measure of  $\Omega$ is defined:

\begin{equation}
\mu_i(m,n)=\sum_{(k,l)\in\Omega} g(k,l).
\end{equation}

\subsection{H\"{o}lder exponent and fine theory}\label{teorfina}

Given a Borel measure $\mu$ and a point $x$ in the support of $\mu$, its local dimension \cite{fal04} in $x$ is

\begin{equation}
    dim_{loc}\mu(x)=\lim_{r\rightarrow0}\frac{\ln(\mu(B(x,r)))}{\ln r}
\end{equation}

if the limit exists, where $B(x,r)$ is a ball of radius $r$ centered in $x$. This limit is usually called H\"{o}lder exponent ($\alpha$). For $\alpha\geq0$ we define

\begin{equation}
    F_{\alpha}=\{x\in spt(\mu):dim_{loc}\mu(x)=\alpha\}.
\end{equation}

We define the Hausdorff fine multifractal spectrum $f_{H}(\alpha)$ as the function that maps $\alpha$ to the Hausdorff dimension $dim_{H}$ of the set $F_{\alpha}$, that is:

\begin{equation}
    f_{H}(\alpha)=dim_{H}F_{\alpha}.
\end{equation}

Although Hausdorff dimension is more convenient for theoretical issues, in practice, it is almost impossible to estimate it from experimental data. That is the reason why the so called 'Coarse Theory' is used to obtain the multifractal spectrum.

\subsection{Coarse theory}\label{teorgruesa}

Once $\alpha$ is calculated for all $x\in spt(\mu)$ as explained in the previous section , $spt(\mu)$ is covered  with a regular lattice of $n$-dimensional boxes of width r (r-mesh). For $r>0$ and $\alpha\geq0$ we define

\begin{equation}
    N_r(\alpha)=\#\{\mbox{ boxes of r-mesh such that } \mu(box)\geq r^{\alpha} \}.
\end{equation}

The coarse multifractal spectrum is defined in \cite{fal04} as

\begin{equation}\label{boxcounting}
    f_C(\alpha)=\lim_{\epsilon\rightarrow0}\lim_{r\rightarrow0}\frac{\ln(N_r(\alpha+\epsilon)-N_r(\alpha-\epsilon))}{-\ln r}
\end{equation}

if the double limit exists.

\subsection{Moment sum and Legendre transformation}\label{metmomentos}

Given $\mu$, its support is covered with an r-mesh. For $q\in \mathbb{R}$ and $r>0$ let us consider the sum of the $q$ powers of the measures $\mu_i$ of each box in the mesh

\begin{equation}
    \chi_q(r)=\sum_{i=1}^{N_r}\mu_i^q,
\end{equation}

where $N_r$ is the total number of boxes in the r-mesh. Consistently with our previous assumptions $\chi_q(r)$ satisfies a power law behaviour with respect to its scale\cite{fal04} at least for small $r$

\begin{equation}\label{leypotmomentos}
    \chi_q(r)\sim r^{-\tau(q)},
\end{equation}

which implies

\begin{equation}\label{legendre}
    \tau(q)=q\alpha+f(\alpha).
\end{equation}

We observe that $\tau(q)$ and $f(\alpha)$  are related through a Legendre transform. If $\tau(q)$ is a convex function, which ensure the Legendre transform is uniquely determined, it is possible to calculate a certain $f_L(\alpha)$ from equation \ref{legendre}. This is known as Legendre spectrum. It can be shown that $f(\alpha)\leq f_L(\alpha)$.

\subsection{Neural networks}

\begin{figure}
        \begin{center}
        \subfigure[Stylized neuron.]{\includegraphics[height=0.2\textheight]{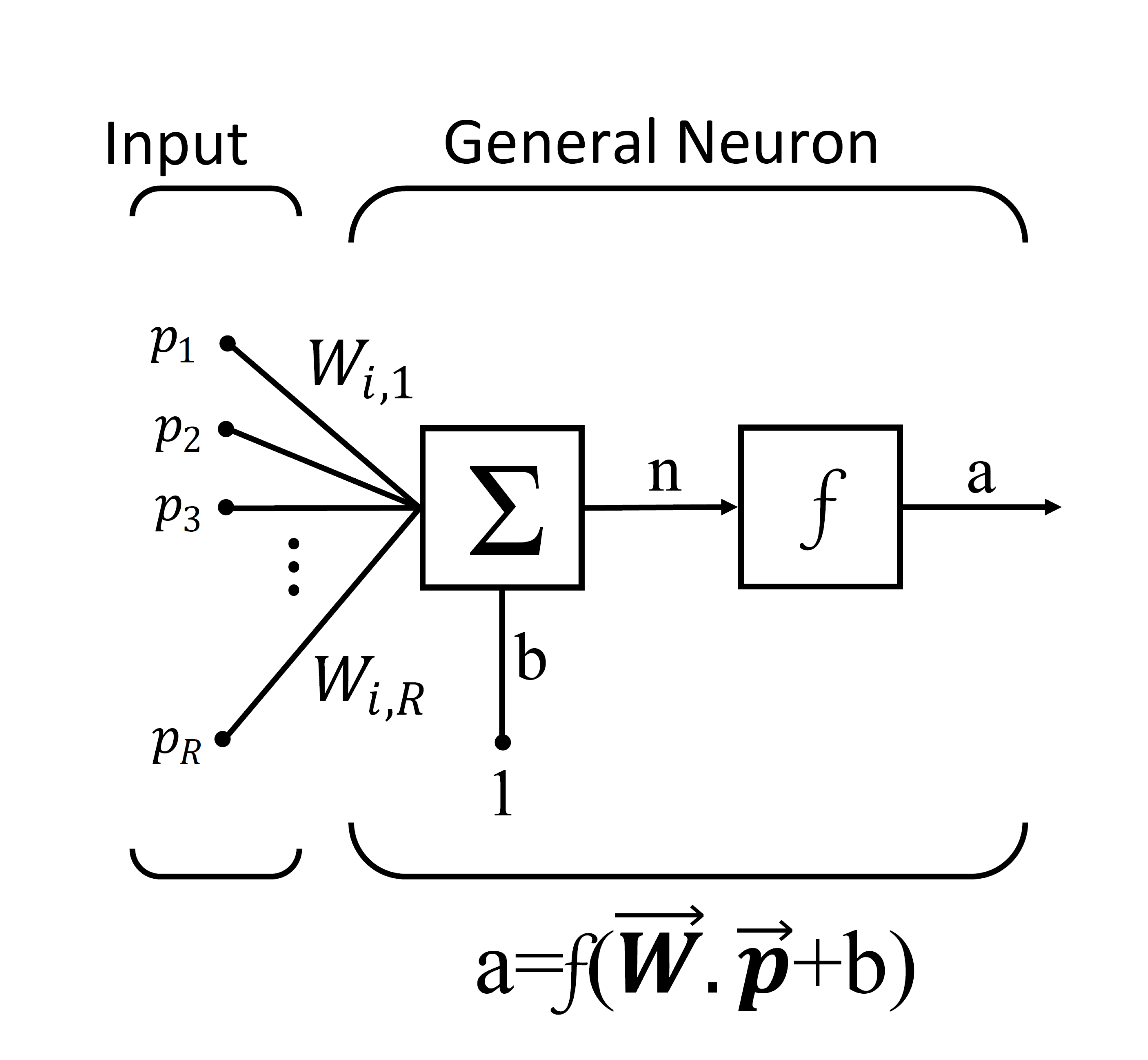} \label{disenoneurona}}
        \subfigure[Feedforward neural network scheme.]{\includegraphics[height=0.2\textheight]{feedforwardeng-eps-converted-to.pdf}\label{feedforward}}
        \caption{Schematic representation of a neural network.}\label{redneuronal}
        \end{center}
\end{figure}

A neuron is a non-linear parameterized function\cite{dre05} that maps the aggregation of many inputs to generate a single output. A neural network is a combination of a given number of neurons, which transfer their results from one to another in a given way. The way in which neurons interconnect and the number of neurons is called the net architecture.

Neural networks are commonly used in various disciplines like robotics, engineering oriented to development of artificial intelligence, remote sensing, medicine, etc., with a wide range of objectives. Among them we can mention classification \cite{kur14,zha14,fon05}, pattern recognition \cite{foo02}, segmentation \cite{bro07,son07}, function approximation, etc.

A schematic representation of a neuron can be seen in figure \ref{disenoneurona}. In general, parameters $\{W_{ij}\}$ are assigned as  weights to the inputs $\{p_j\}$ of the neuron. The output of the neuron is a non-linear function $f$ of the sum of its inputs weighted by parameters. A constant bias is added. $f$ is called activation function and  it is usually a sigmoid function.

There are several neural network architectures; particularly, a feedforward neural network is an arrangement of neurons in two or more layers, in which the outputs of one layer are the inputs of the next one, (see figure \ref{feedforward}).

\begin{figure}
\centering
\begin{minipage}[c]{.39\linewidth}
\centering
      \includegraphics[width=\textwidth]{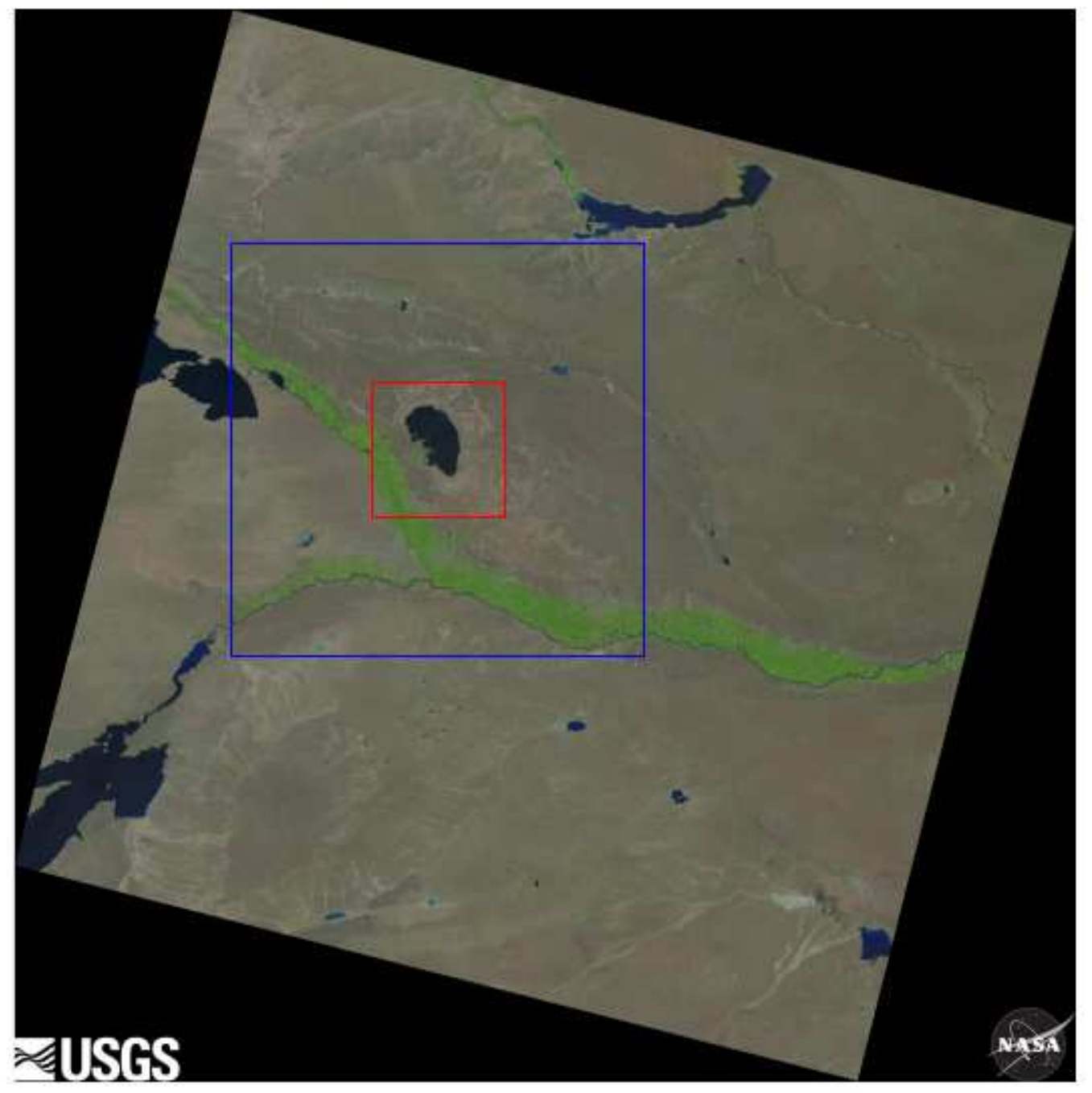}\\
      \vspace{-1.8cm}
      {\scriptsize
\begin{flushleft}
Description:
\end{flushleft}

 \begin{tabular}{|l|c|}
  \hline
  Designation: & LC8230087 \\
  \hline
  Acquisition date: & 08/02/2015 \\
  \hline
  WRS-2 Path: & 230 \\
  \hline
  WRS-2 Row: & 087 \\
  \hline
  Latitude: & 38º 54' S \\
  \hline
  Longitude: & 67º 45' \textmd{O}\\
  \hline
  Spacial resolution (m): & 30 \\
  \hline
\end{tabular}
\begin{flushleft}
{\color{red}{\large -}} Analyzed region

{\color{blue}{\large -}} Padding
\end{flushleft}
}
\end{minipage}
\begin{minipage}[c]{.39\linewidth}
\centering
      \includegraphics[width=\textwidth]{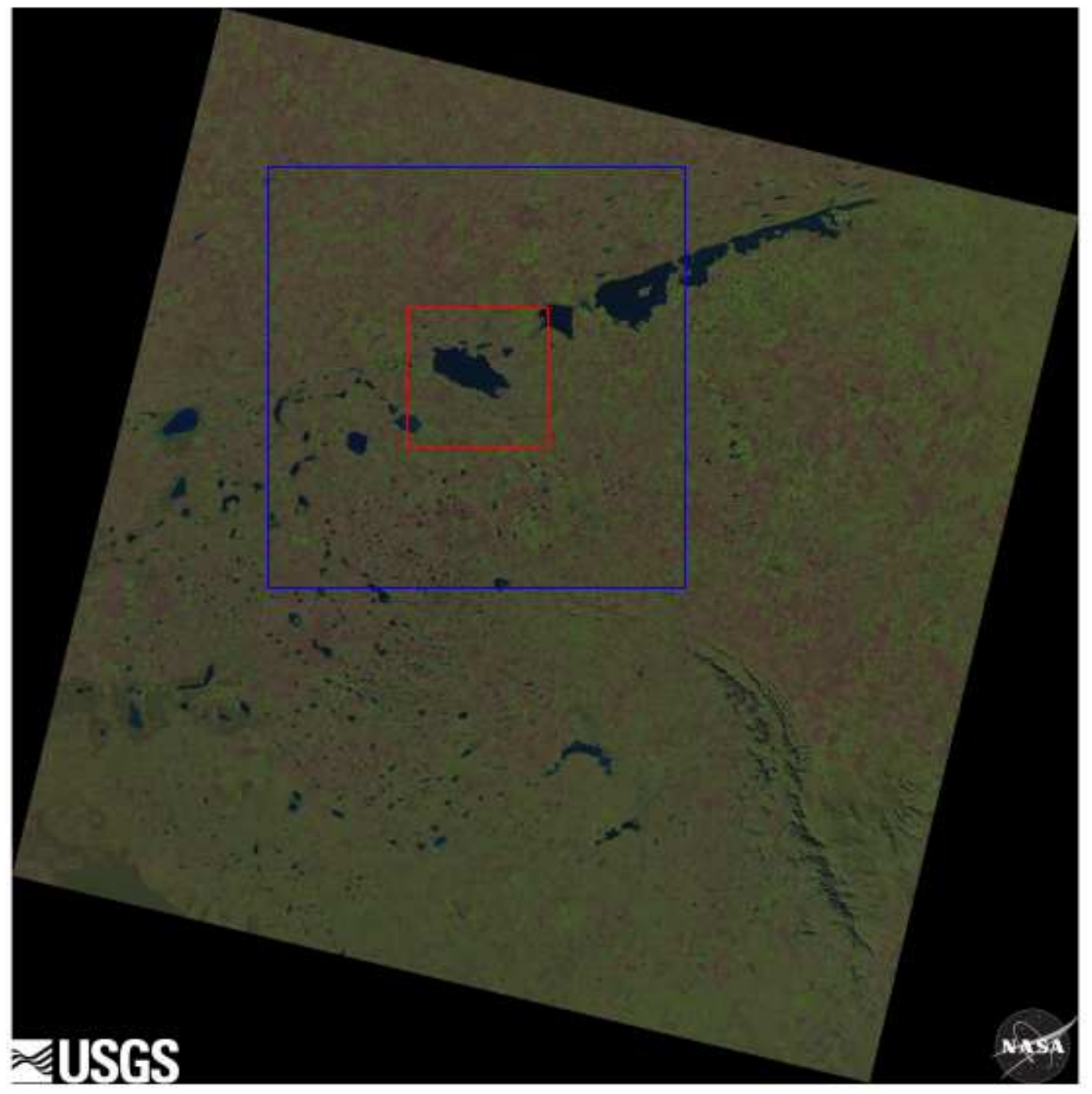}\\
      \vspace{-1.8cm}
      {\scriptsize
\begin{flushleft}
Description:
\end{flushleft}
 \begin{tabular}{|l|c|}
  \hline
  Designation: & LC8227086 \\
  \hline
  Acquisition date: & 11/06/2015 \\
  \hline
  WRS-2 Path: & 227 \\
  \hline
  WRS-2 Row: & 086 \\
  \hline
  Latitude: & 37º 29' S \\
  \hline
  Longitude: & 62º 41' \textmd{O} \\
  \hline
  Spacial resolution (m): & 30 \\
  \hline
\end{tabular}
\begin{flushleft}
{\color{red}{\large -}} Analyzed region

{\color{blue}{\large -}} Padding
\end{flushleft}
}
\end{minipage}\\
  \caption{True color optical images used in the analysis. The red (inner) squares delimit the regions fully analyzed. The blue (outer) squares delimit the padding regions}\label{opticas}
\end{figure}

\begin{figure}
\begin{minipage}[c]{.3\linewidth}
\subfigure[Landsat image used in the analysis.]{\includegraphics[width=\textwidth]{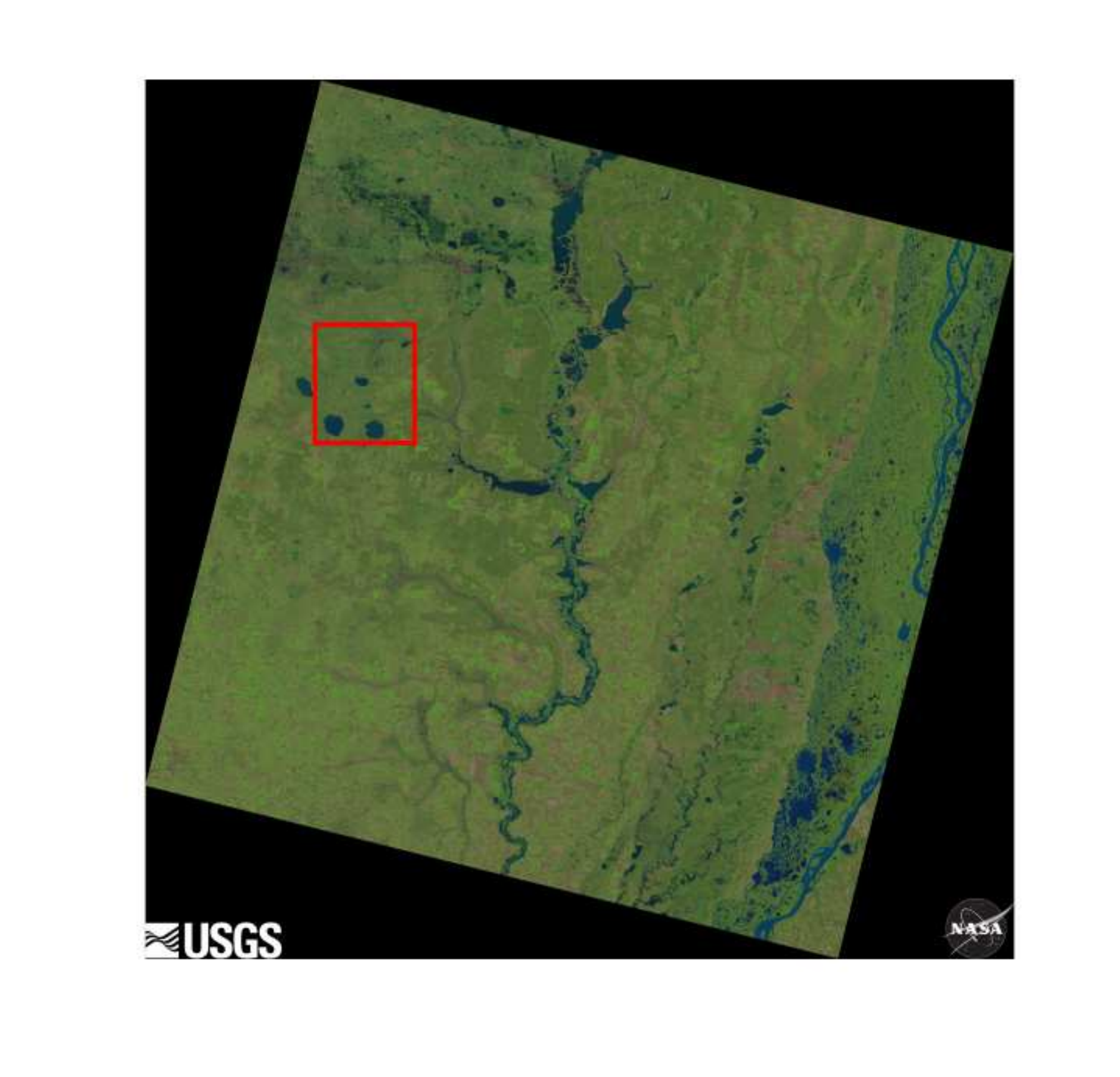}\label{S1referencia005886}}
\end{minipage}
\begin{minipage}[c]{.3\linewidth}
\subfigure[False-Color Composite SAR image of a subregion.]{\includegraphics[width=\textwidth]{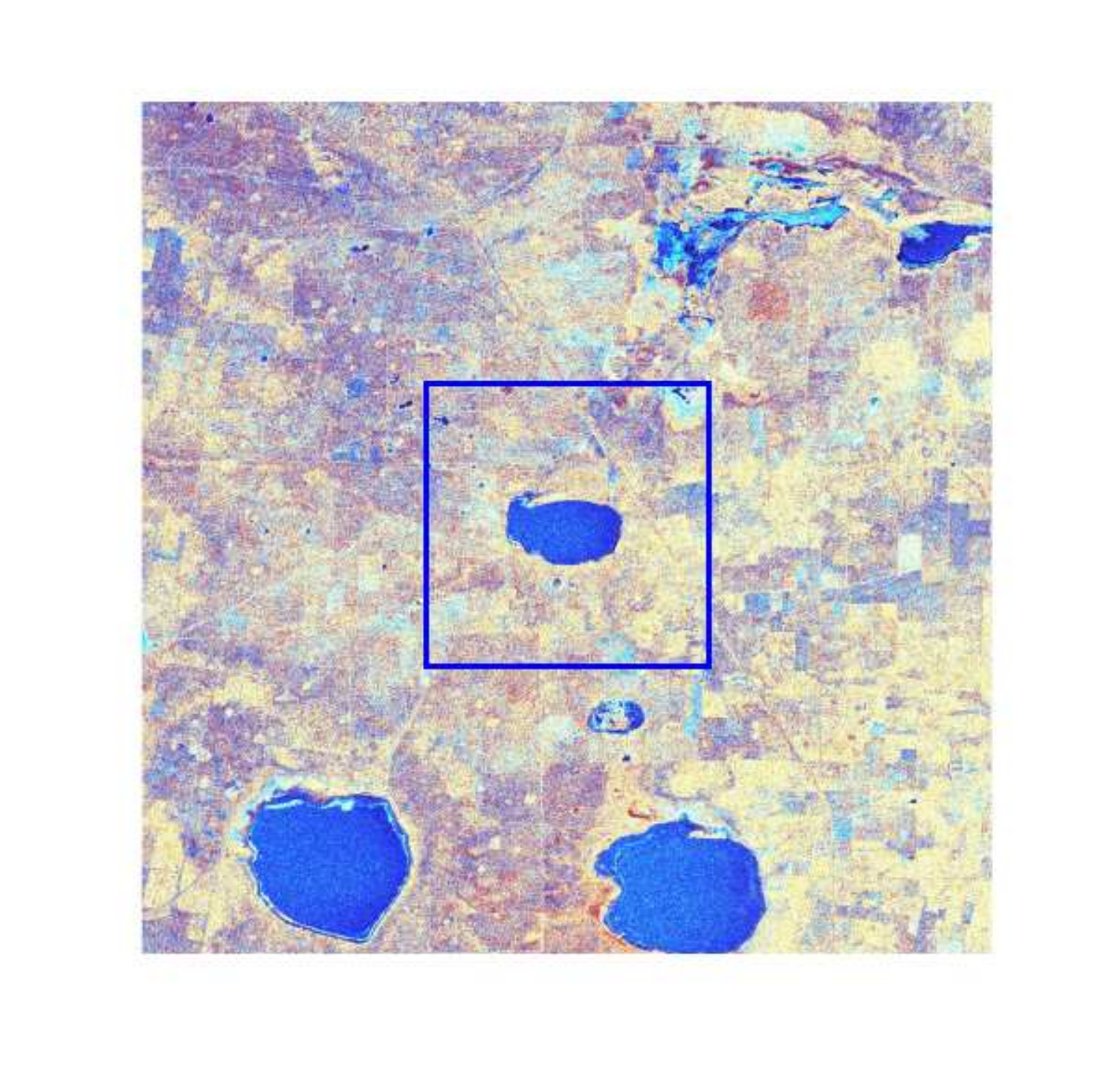}\label{S1falsocolor005886}}
\end{minipage}
\begin{minipage}[c]{.25\linewidth}
{
\scriptsize
 Description:

 \scalebox{0.8}{\begin{tabular}{|l|c|}
  \hline
  Designation: & S1005886 \\
  \hline
  Product type: & Level 1 GRD (High resolution) \\
  \hline
  Sensor Mode: & Interferometric Wide swath \\
  \hline
  Polarization & Dual (VV, VH) \\
  \hline
  Acquisition date : & 12/05/2015 \\
  \hline
  WRS-2 Path: & 227 \\
  \hline
  WRS-2 Row: & 081 \\
  \hline
  Latitude: & 30º 18'S \\
  \hline
  Longitude: & 60º 41'\textmd{O} \\
  \hline
  Spacial resolution (m): & 10 \\
  \hline
\end{tabular}}

{\color{blue}{\large -}} Analyzed region

{\color{red}{\large -}} Padding}
\end{minipage}
\caption{Description of image S1005886.}\label{presentacionS1005886}
\end{figure}

\begin{figure}
\begin{minipage}[c]{.3\linewidth}
\subfigure[Landsat image used in the analysis.]{\includegraphics[width=\textwidth]{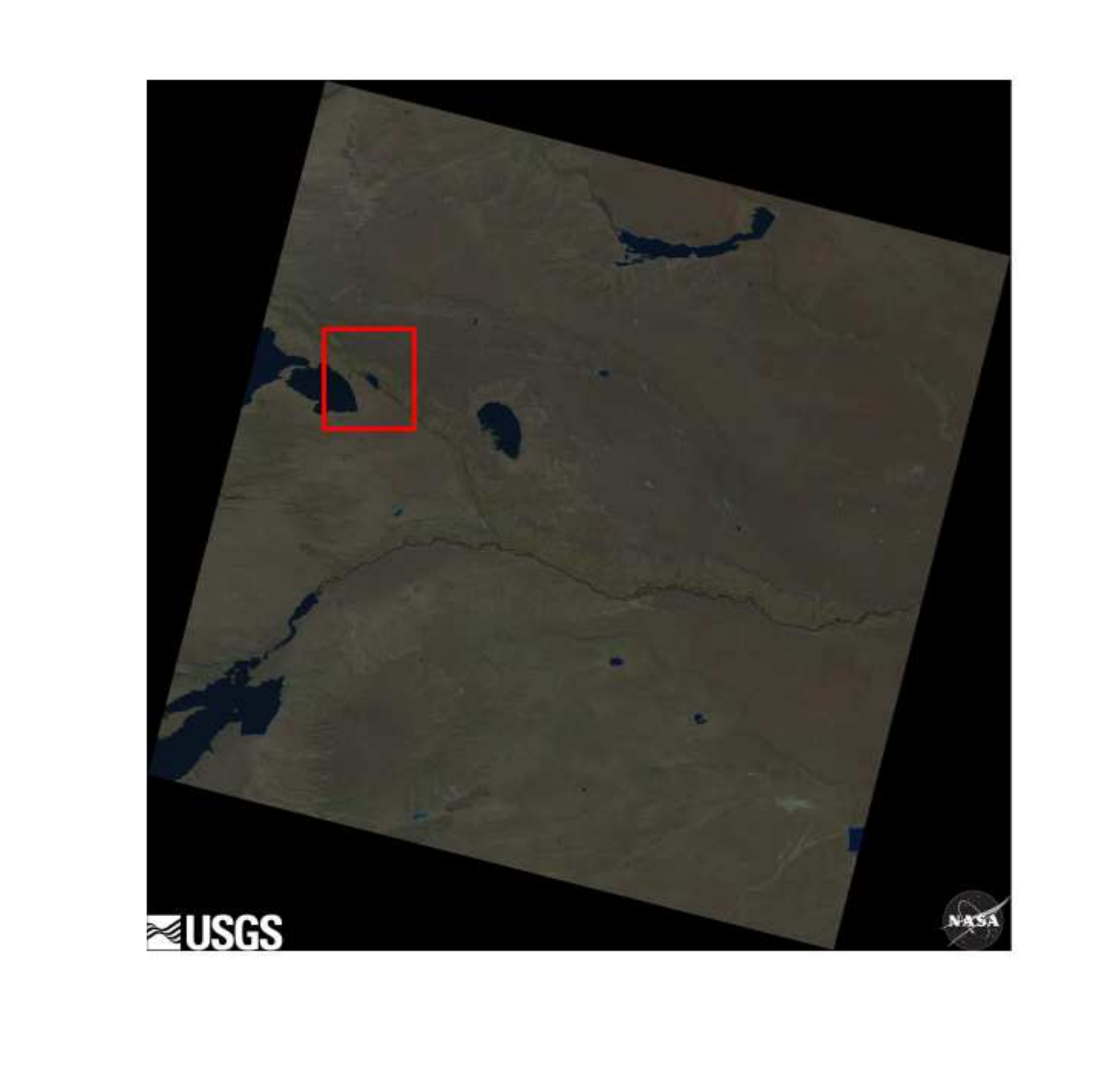}\label{S1referencia005959}}
\end{minipage}
\begin{minipage}[c]{.3\linewidth}
\subfigure[False-Color Composite SAR image from a subregion.]{\includegraphics[width=\textwidth]{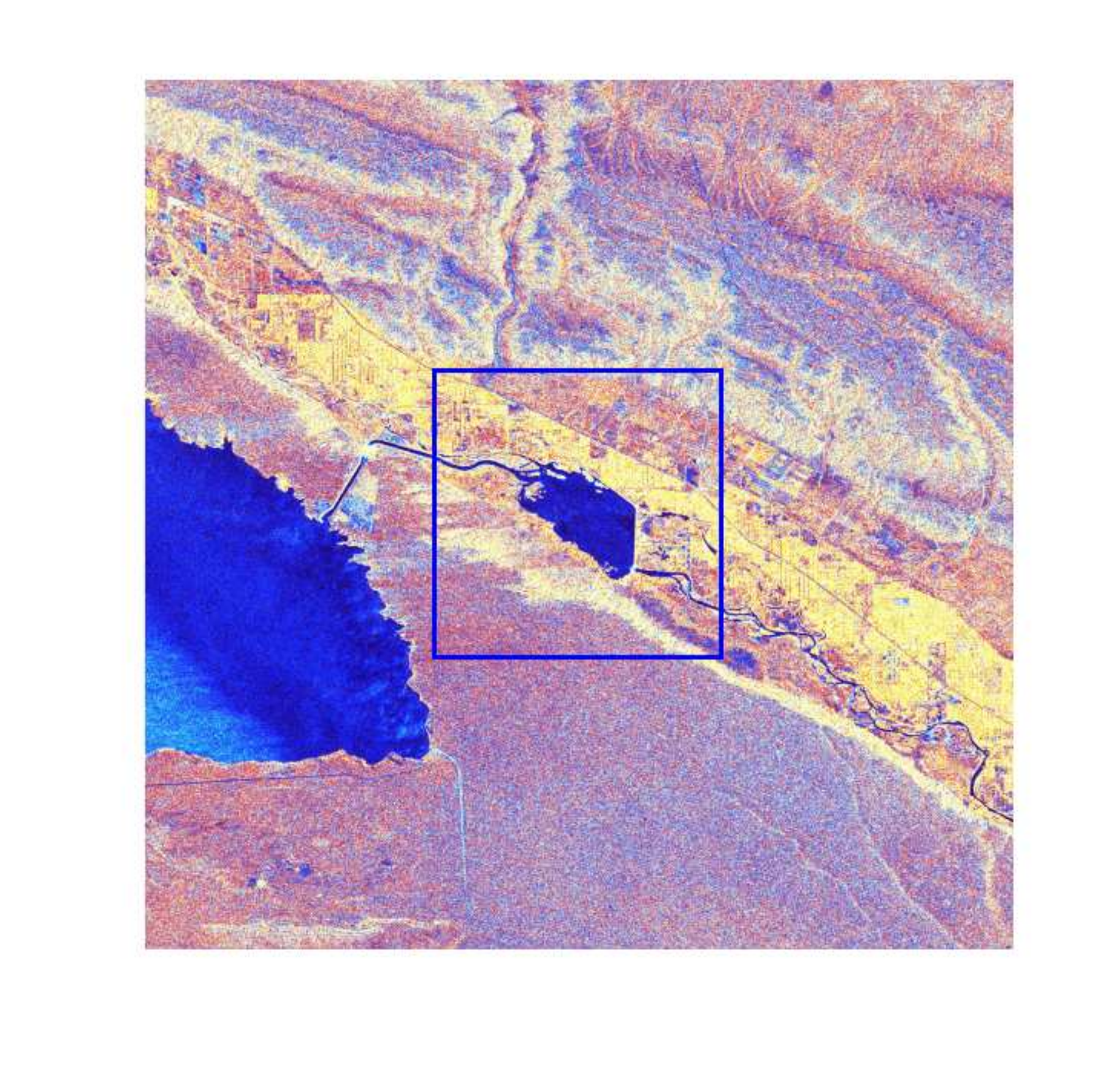}\label{S1falsocolor005959}}
\end{minipage}
\begin{minipage}[c]{.25\linewidth}
{
\scriptsize
 Description:

 \scalebox{0.8}{\begin{tabular}{|l|c|}
  \hline
  Designation: & S1005959 \\
  \hline
  Product type: & Level 1 GRD (High resolution) \\
  \hline
  Sensor Mode: & Interferometric Wide swath \\
  \hline
  Polarization & Dual (VV, VH) \\
  \hline
  Acquisition date: & 17/05/2015 \\
  \hline
  WRS-2 Path: & 230 \\
  \hline
  WRS-2 Row: & 087 \\
  \hline
  Latitude: & 38º 54'S \\
  \hline
  Longitude: & 67º 45'\textmd{O} \\
  \hline
  Spacial resolution (m): & 10 \\
  \hline
\end{tabular}}

{\color{blue}{\large -}} Analyzed region

{\color{red}{\large -}} Padding}
\end{minipage}
\caption{Description of image S1005959.}\label{presentacionS1005959}
\end{figure}

\section{Description of methods}\label{metodologia}

The analysis performed on images is detailed in this section. In subsection \ref{preprocesamiento}, we present two optical and two SAR images which will be used as examples of application of the multifractal method and summarize its preprocessing. In subsection \ref{amultifractal} we discuss how the tools of multifractal theory, presented in subsections \ref{medidas} - \ref{metmomentos}, are used to obtain a classification of optical and radar images, clarifying the differences between these cases. Subsection \ref{aneuronales} details the training and use of neural networks for land cover classification and the implemented architectures for each type of image. Finally, the use of NDWI for obtaining the region covered by water in the case of optical images is summarized in subsection \ref{andwi}.

\subsection{Analyzed images}\label{preprocesamiento}

Optical images taken by OLI (Operational Land Imager) and TIRS (Thermal Infrared Sensor) sensors of Landsat 8 satellite, provided by the U.S. Geological Survey\cite{ee}, were studied. Two examples are presented in figure \ref{opticas}. For more information on the Landsat 8 platform we refer to \cite{lan15}.

For each analized image, we select and cut a subset of $3072\times3072$ pixeles. Next, the bands of the subsets are calibrated so that their values represent reflectance measured by the sensor. Finally, bands corresponding to the frequencies of blue, green, red and near infrared are selected.

We also worked with synthetic aperture radar (SAR) images of the Sentinel-1 platform of the European Space Agency (ESA) \cite{hub}. A complete description of the operation of the satellite Sentinel-1 can be found in \cite{sen13}.

SAR images analyzed in this paper are already projected to ground range.  They have dual polarizations, so they are composed of two bands (VV, VH).

Images are calibrated transforming their digital values into backscattering coefficient values ($\sigma_0$). Finally, they are cut, obtaining images with two bands corresponding to VH and VV polarizations of $3072\times3072$ pixels.

The Sentinel-1 images analyzed in this work are described in figures \ref{presentacionS1005886} and \ref{presentacionS1005959}. In them we see the subsets studied in a false colour composite where information of the bands ($\sigma_0$-VH), ($\sigma_0$-VV) and ($\sigma_0$-VV)/($\sigma_0$-VH) was placed in the red, green and blue channels, respectively. The regions limited by a blue box are the regions fully analyzed with our methods, while the remaining fragments were used as padding for the calculation of the regularity of the pixels on the border.

\subsection{Analysis through the multifractal formalism}\label{amultifractal}

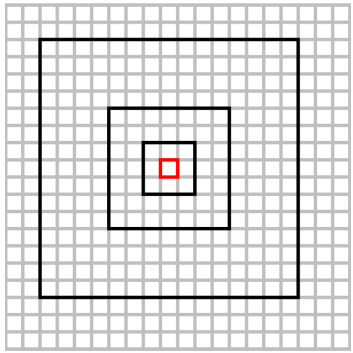
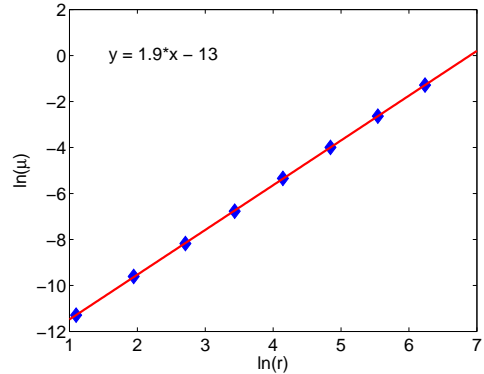
\begin{figure}
\begin{minipage}{.4\linewidth}
            \begin{center}
            \subfigure[{\footnotesize Examples of the used neighbourhoods:\newline{\color{red}{\large -}} Pixel whose $\alpha$ is calculated;\newline{\color{black}{\large -}} Some possible neighbourhoods.}]{\includegraphics[height=0.7\linewidth]{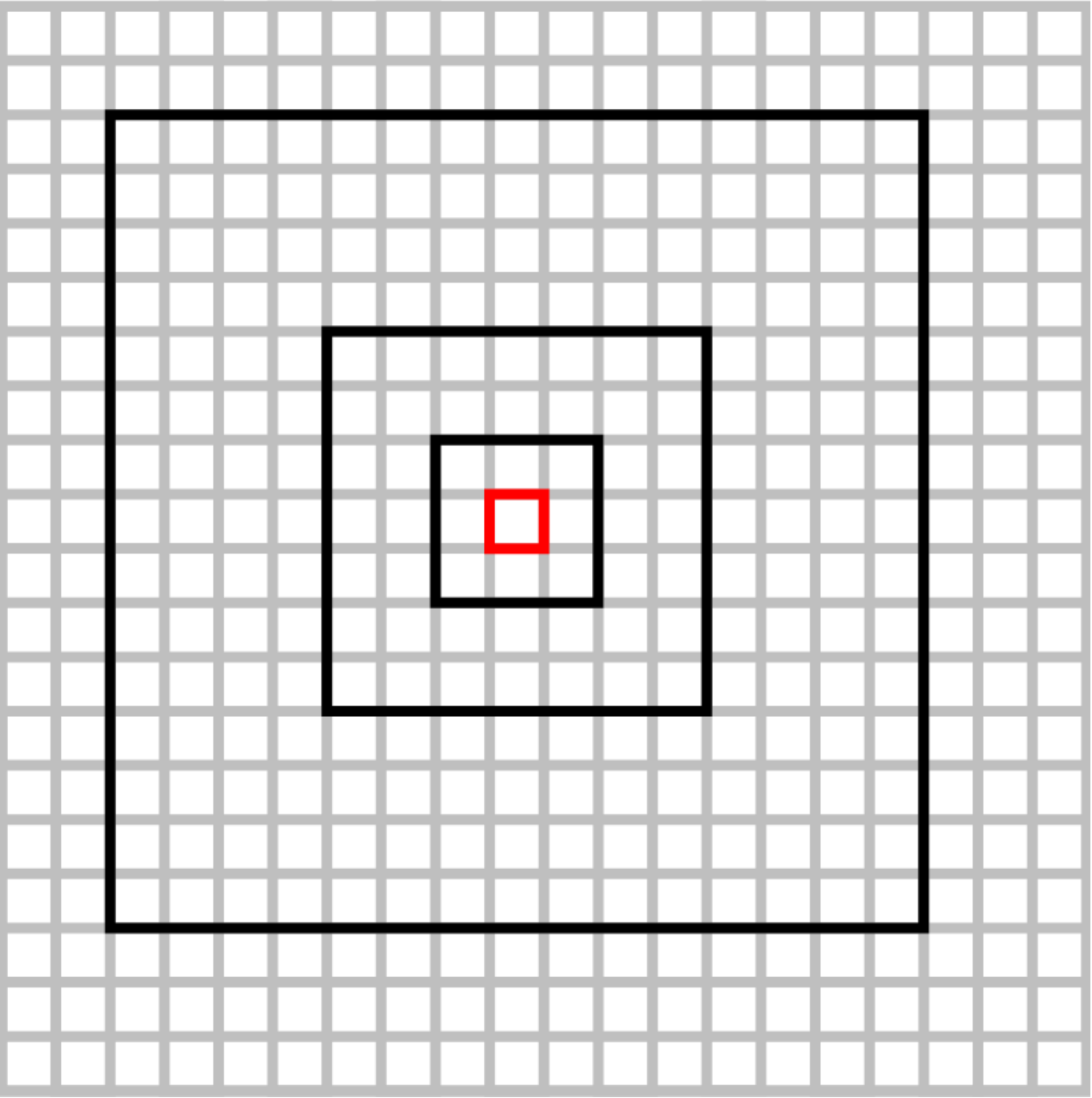}\label{rlocalentornos}}
            \end{center}
            \end{minipage}
            \begin{minipage}{.15\linewidth}
                \begin{tikzpicture}
                \draw[black, -latex, line width=8pt] (0,2) -- (2.1,2);
                \draw[blue, -latex, line width=7pt] (0.01,2) -- (2,2);
                \draw[white, -, line width=0.1pt] (0,0) -- (2,0);
                \end{tikzpicture}
            \end{minipage}
            \begin{minipage}{.4\linewidth}
            \begin{center}
            \vspace{-1cm}
            \subfigure[Linear fit.]{
                \includegraphics[height=0.8\linewidth]{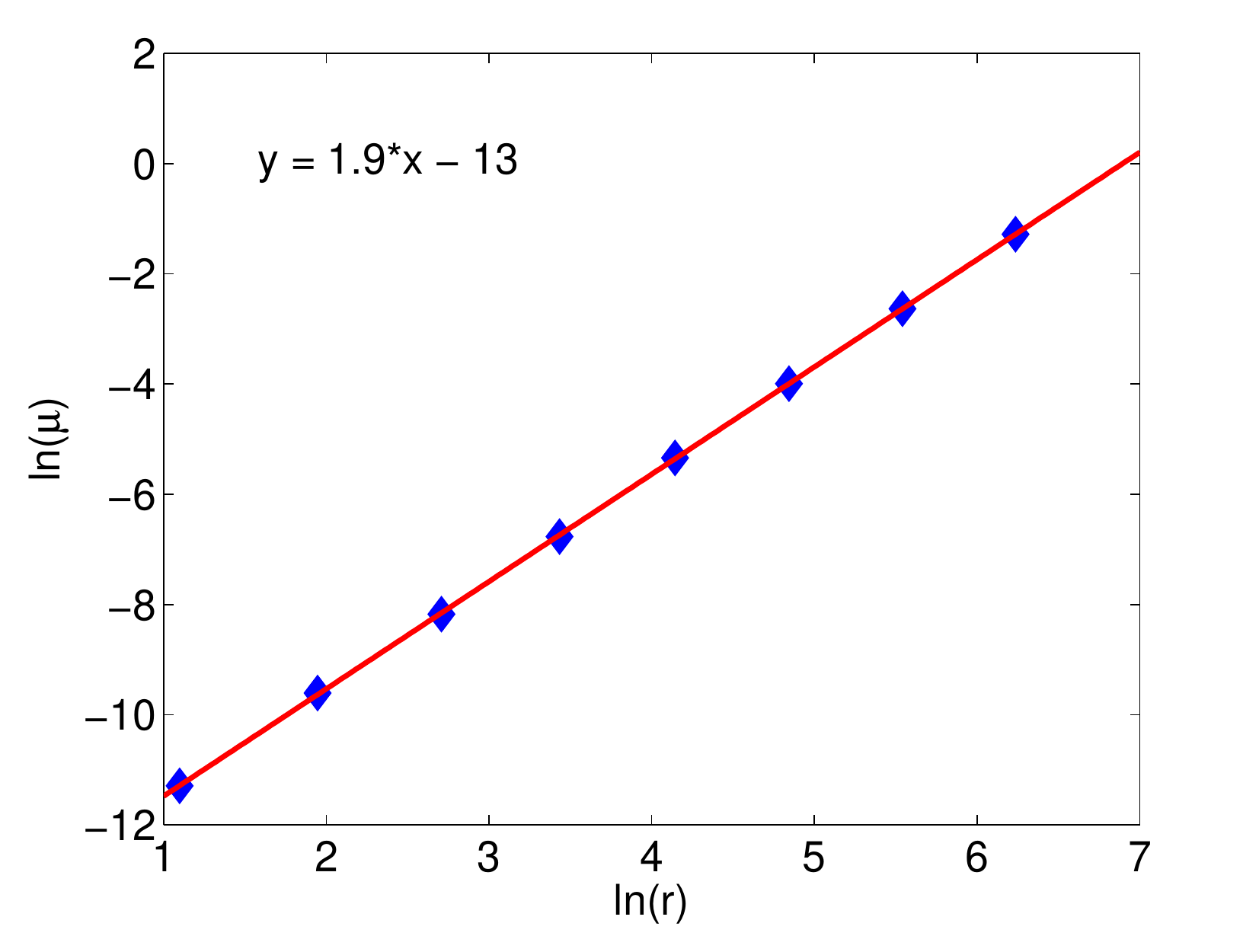}\label{rlocalajuste}}
                \bigskip
            \end{center}
            \end{minipage}\\
  \caption{Calculation of $\alpha$ at a pixel: $\mu$ is measured on a series of neighbourhoods with different widths $r$ (those with $r=3,7,15$ are presented in \ref{rlocalentornos}). A linear fit is done over the data pairs $(\ln(r), \ln(\mu))$, (figure \ref{rlocalajuste}). The slope of the fit is the $\alpha$ corresponding to that pixel.}\label{rlocalcalculo}
\end{figure}

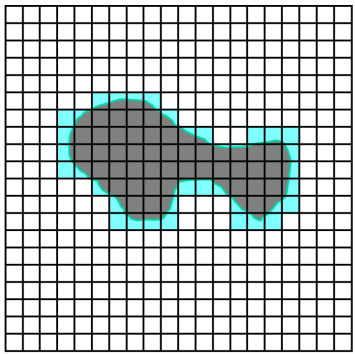
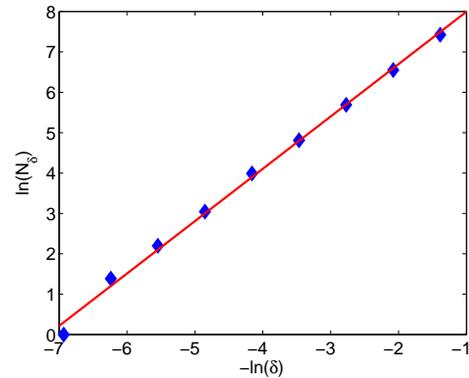
\begin{figure}
\begin{minipage}{.4\linewidth}
            \begin{center}
                \subfigure[Set of pixels of interest covered by boxes of width $\delta$. The number of boxes needed to cover the whole set is $N_\delta$.]{\includegraphics[height=0.7\linewidth]{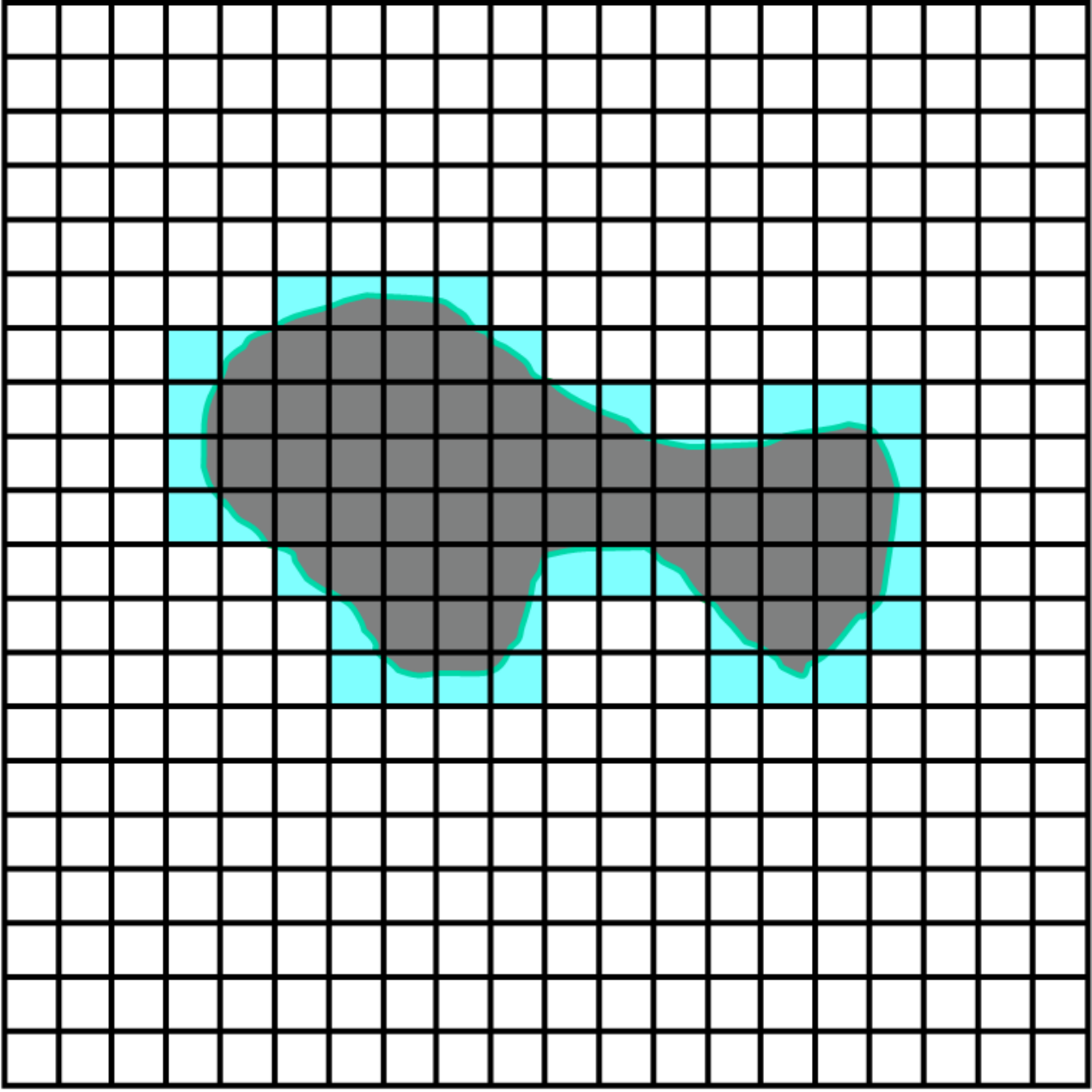}\label{dglobalmalla}}\\
            \end{center}
            \end{minipage}
            \begin{minipage}{.15\linewidth}
                \begin{tikzpicture}
                \draw[black, -latex, line width=8pt] (0,2) -- (2.1,2);
                \draw[blue, -latex, line width=7pt] (0.01,2) -- (2,2);
                \draw[white, -, line width=0.1pt] (0,0) -- (2,0);
                \end{tikzpicture}
            \end{minipage}
            \begin{minipage}{.4\linewidth}
            \begin{center}
            \vspace{-1.5cm}
                \subfigure[Linear fit.]{\includegraphics[height=0.8\linewidth]{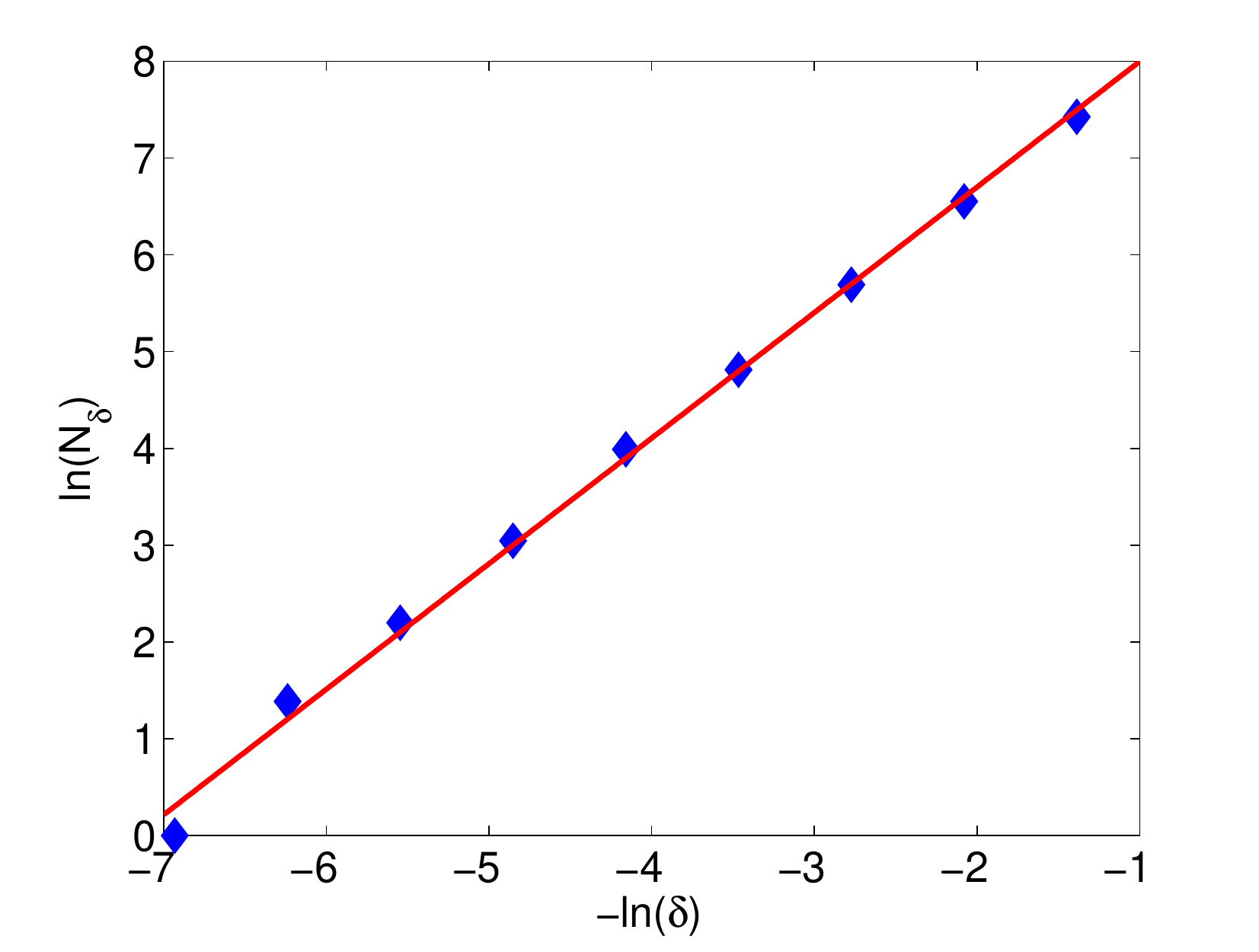}}\\
            \end{center}
            \end{minipage}\\
  \caption{Calculation of $f(\alpha)$: the image is covered with meshes of several box widths and the number of boxes needed to cover the set of pixels of interest is counted (figure \ref{dglobalmalla}). The data pairs $(\ln(N_\delta),-\ln(\delta))$ are adjusted with a linear function. The slope of the adjusted function is the fractal dimension  $f(\alpha)$.}\label{dglobalcalculo}
\end{figure}

This analysis is applied on the $1024\times1024$ pixels of the central region of the near infrared band in the Landsat images (figure \ref{opticas}). It is also applied on the same region of the $\sigma_0$ VH bands in the Sentinel-1 images (figures \ref{presentacionS1005886} y \ref{presentacionS1005959}).

We calculate $\alpha$ for every pixel of the analyzed region with the measure $\mu$ defined earlier. We choose the $B(x,r)$ in equation \ref{boxcounting} to be square neighbourhoods of $(2k-1)\times(2k-1)$ pixels, with $k=2,\ldots,9$ for the optical images and $k=3,\ldots, 9$ for the radar images. Those neighbourhoods are centered in the pixel whose H\"{o}lder exponent we are trying to calculate. For SAR images, the neighbourhoods corresponding to $k=1$ and $k=2$ were discarded because they are strongly affected by speckle noise. The normalized sum of the pixel values is calculated for all neighbourhoods. A linear fit is then performed on the logarithms of the measures with respect to the logarithms of the neighbourhood widths (figure \ref{rlocalcalculo}). The slope of the adjusted line is our estimate of $\alpha$ for that pixel. The calculations on the pixels near the border of the analyzed region are made using adjacent regions of the image as padding.

Once $\alpha$ values for every pixel are determined, the range of $\alpha$ is divided into R=30 classes $[\alpha_{min}+(s-1)\cdot \delta \alpha,\alpha_{min}+ s\cdot\delta \alpha]$, with $\delta \alpha =(\alpha_{max}- \alpha_{min})/R$ and $s=1,\ldots,30$. Those classes determine subsets of pixels with similar local regularity. We then proceed to calculate the box counting dimension of each subset. For this, different meshes are applied to the image and the number of boxes necessary to completely cover the selected subset is counted. For each subset we make a linear fit of the logarithm of the number of boxes with respect to minus the logarithm of the width of the boxes (figure \ref{dglobalcalculo}). The slope of the fit is our estimate of $f(\alpha_{m})$, where $\alpha_{m}$ is the mean value of each class. The boxes of the meshes used in both types of images have $4, 8, 16, 32, 64, 128, 256, 512$ and $1024$ pixels of width.

In order to get more detail on $f(\alpha)$  in the regions where $\alpha$ is near to $\alpha_{min}$ or $\alpha_{max}$, we divide the first and last classes in halves. We calculate the box counting dimension for the subsets determined by the first half of the $s=1$ class and the last half of $s=30$ class. The values obtained are associated to $f(\alpha_{min})$ and $f(\alpha_{max})$ respectively.

A polynomial is fitted to the multifractal spectrum and the corresponding value of $f(\alpha)$ is associated to each pixel of the image.

We have also determined the multifractal spectrum through a Legendre transform. Different meshes with boxes  $ 4, 8, 16, 32, 64, 128, 256, 512$ and $ 1024 $ pixels wide are applied over the image. Given a mesh, the sum of the reflectances of the pixels within each box is calculated. These measures are raised to an exponent $q$ and added, thus obtaining the value of the partition function for that mesh width and the exponent $q$. After making this calculation for each mesh, a linear fit of the logarithm of the partition functions with respect to the negative of the logarithms of the corresponding box widths is performed. This gives us $\tau(q)$ of equation \ref{leypotmomentos} for each $q$. After applying a Legendre transform to $\tau (q)$ a statistical estimation of the multifractal spectrum is obtained. Although this does not give us local information on the regularity of each pixel as coarse theory does, it is computationally faster and allows us to achieve an overall approximation of the behaviour of the spectrum.

The classification of the pixels of the images in flooded or not flooded is achieved through the application of thresholds on their multifractal spectra. The selection of these thresholds arose during the exploratory study of the graphs of $f(\alpha)$ obtained and it will be detailed in section \ref{resultados}.

\subsection{Analysis using neural networks}\label{aneuronales}

Feedforward neural networks were trained to perform segmentation of water bodies in the images. Landsat images were analyzed with a network with 4 neurons in the input layer (corresponding to each of the bands of the images: red, green, blue and near infrared), a hidden layer with 20 neurons and 2 neurons in the output layer (associated with the two categories of cover: water and not water). The analysis of the SAR images was done through a network of two input neurons (associated with two bands of information), a hidden layer with 30 neurons, and two output neurons. All neurons have a sigmoid function response.

Neural networks were trained through the scaled conjugate gradient backpropagation method \cite{mol93}. For each type of images, training regions were selected outside the studied regions. $70\%$ of training data is used to train the network, $15\%$ is reserved to validate the result and the remaining $15\%$ is used to test the network performance. The selection of data for each group is performed randomly.

\begin{figure}
\centering
  \subfigure[Classification.]{\includegraphics[width=0.4\textwidth]{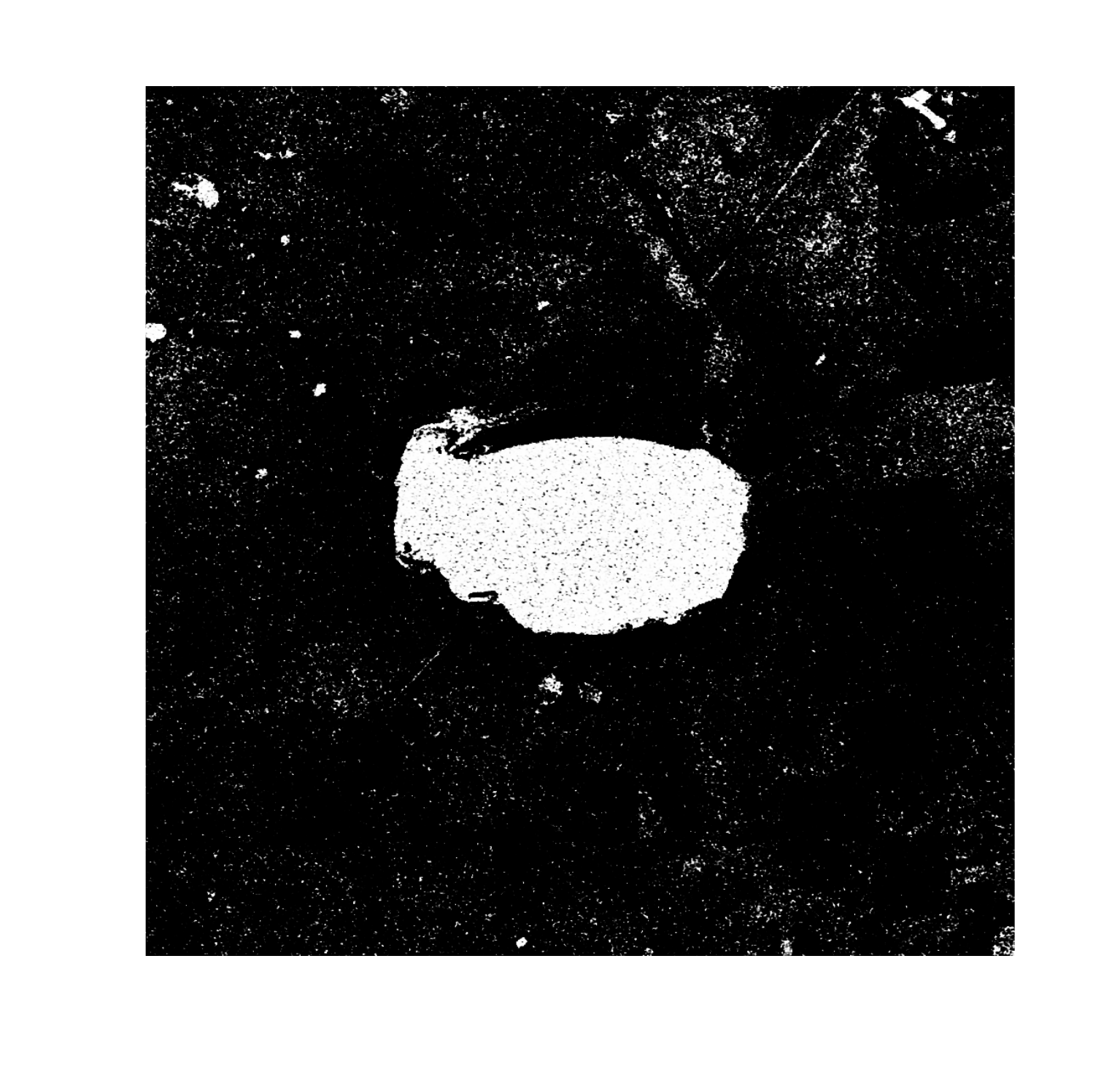}\label{neuronasegmentacionpuraS1005886}}
  \subfigure[Classification filtered.]{\includegraphics[width=0.4\textwidth]{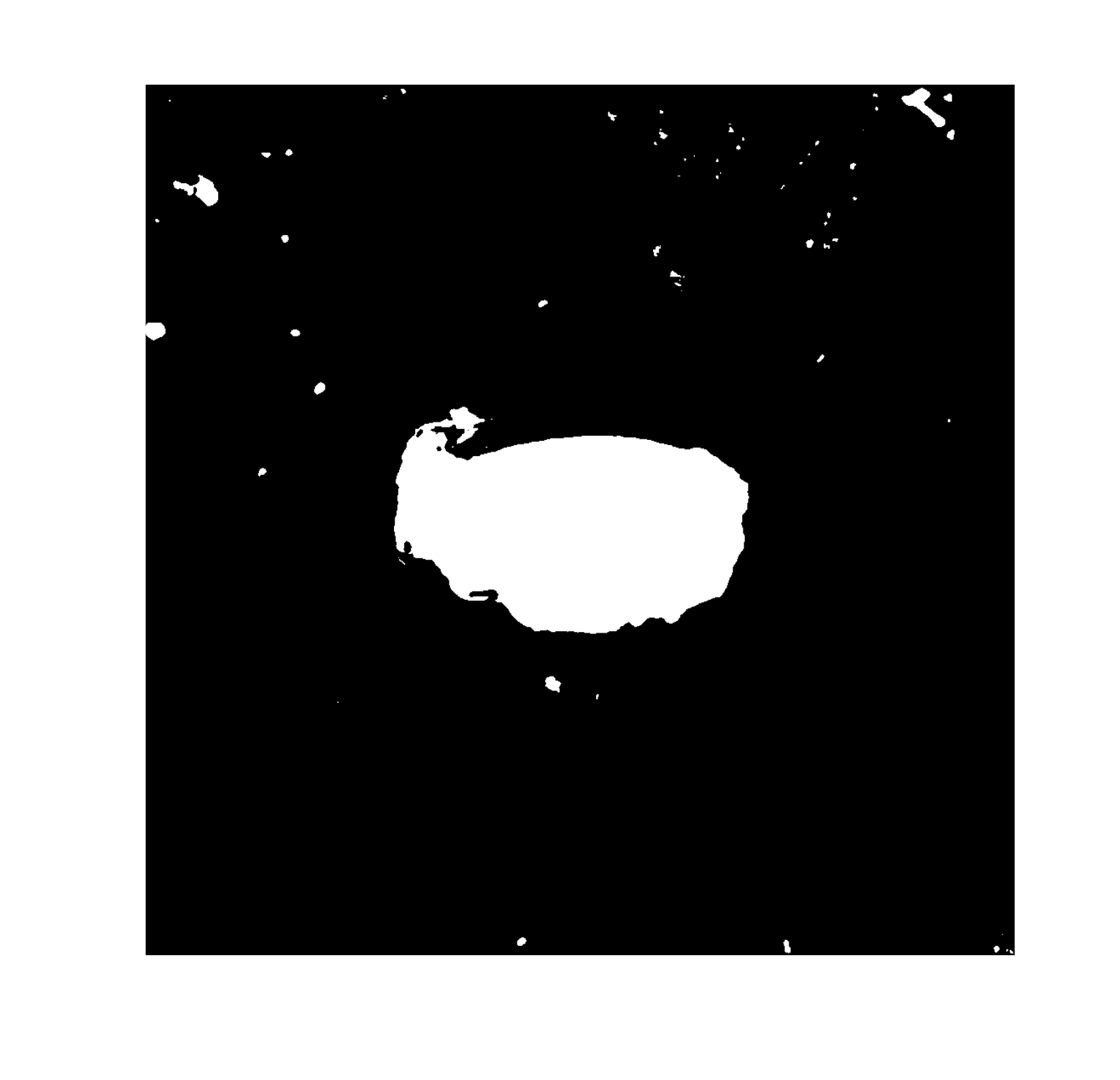}\label{neuronasegmentacionfiltradaS1005886}}
  \caption{Classification made with the neural network over the image S1005886.}\label{neuronassegmenacionS1005886}
\end{figure}

We observe that the classification resulting from the network used on radar images does not met fully the expectations. For example, in figure \ref{neuronasegmentacionpuraS1005886} there are pixels classified as non-water in regions where there is water. This is probably due to the amount of speckle noise affecting these images, plus the too small number of bands in the input. To improve on this result, other net architectures were tested but with no significant improvement in the classification. It was then decided to keep the previous one and to apply a filter to the output of the network. It has a square kernel of $7\times7$ pixels which gives the central pixel the value of the class that possesses greater proportion within the neighbourhood (figure \ref{neuronasegmentacionfiltradaS1005886}).

\subsection{NDWI}\label{andwi}

The Normalized Differential Water Index proposed in \cite{rog04} is calculated for the Landsat images. We can assume that a pixel is covered by a linear combination of water and other covers, considering it flooded if at least $50\%$ of its surface is covered by water. Let $p_a$ be the proportion of water of the pixel and let $p_{i}$ be the proportion of other covers, with

\begin{equation}
    1=p_a+\sum_i p_{i}
\end{equation}

The reflectance of the pixel is

\begin{equation}
    \rho^k=p_a \rho_{a}^k + \sum_i p_{i} \rho_{i}^k
\end{equation}

where $\rho_a^k$ and $\rho_{i}^k$ are water reflectance and other covers reflectance in the band $k$.

In \cite{rog04} NDWI is defined as

\begin{equation}
    NDWI=\frac{(\rho^r-\rho^{swir})}{(\rho^r+\rho^{swir})}
\end{equation}

where $k=r$ refers to the frequency band of red (band 4 in Landsat 8 images), and $k=swir$  refers to the band of shortwave infrared (band 6).

Therefore,

\begin{equation}\label{NDWI}
    NDWI=\frac{p_a(\rho^r_a-\rho^{swir}_a)+\sum_i p_{i}(\rho^r_{i}-\rho^{swir}_{i})}{(\rho^r+\rho^{swir})}
\end{equation}

An analysis of typical spectral signatures shows that $(\rho^r_{i}-\rho^{swir}_{i})<0$ and $(\rho^r_a-\rho^{swir}_a)>0$.

As the NDWI takes values in $[-1,1]$, when $p_a\geq0.5$, from \ref{NDWI} it follows that the $NDWI\geq0$. This is the threshold used in the classification.

\section{Results and Discussion}\label{resultados}

\begin{figure}
\centering
        \subfigure[Comparison between multifractal spectra obtained by the coarse theory and the statistical method for the image LC8230087.]{\includegraphics[width=0.48\textwidth]{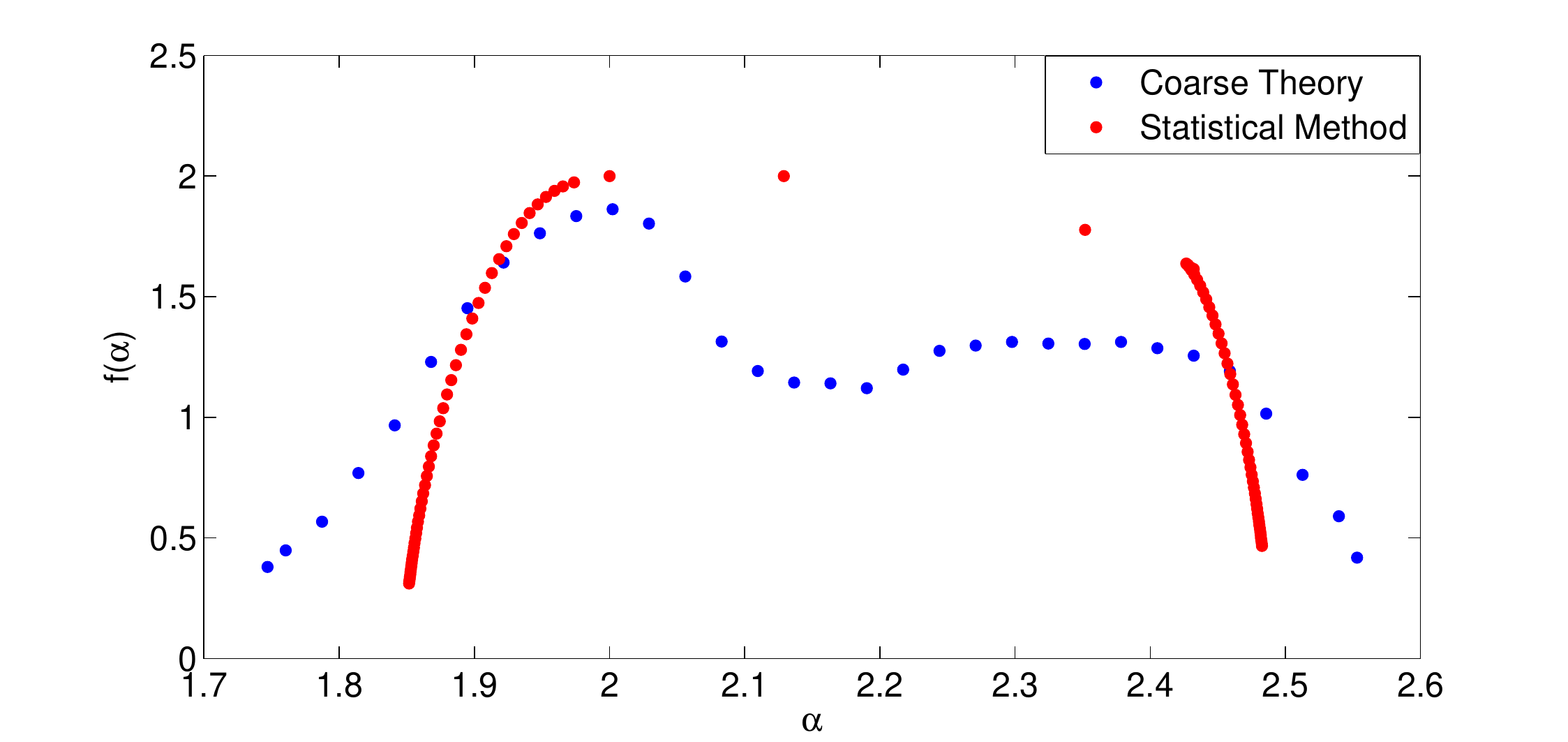}\label{fractalcomparacionLC8230087}}\hfill
        \subfigure[Comparison between multifractal spectra obtained by the coarse theory and the statistical method for the image S1005959.]{\includegraphics[width=0.48\textwidth]{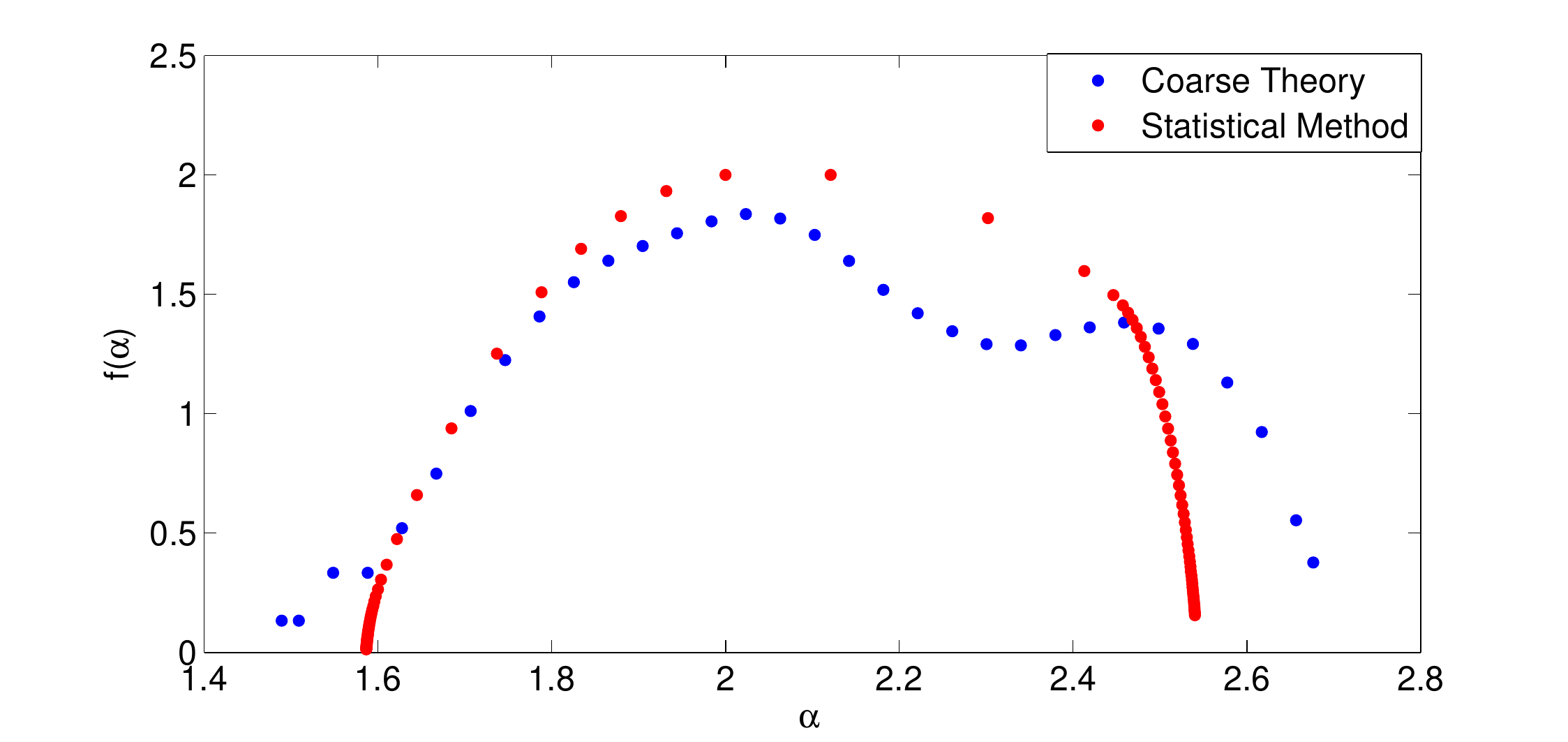}\label{fractalcomparacionS1005959}}\\
\caption{Comparison between multifractal spectra obtained by the coarse theory and the statistical method.}
\label{fractalcomparacion}
\end{figure}

\begin{figure}
\centering
        \subfigure[True color image.]{\includegraphics[width=0.4\textwidth]{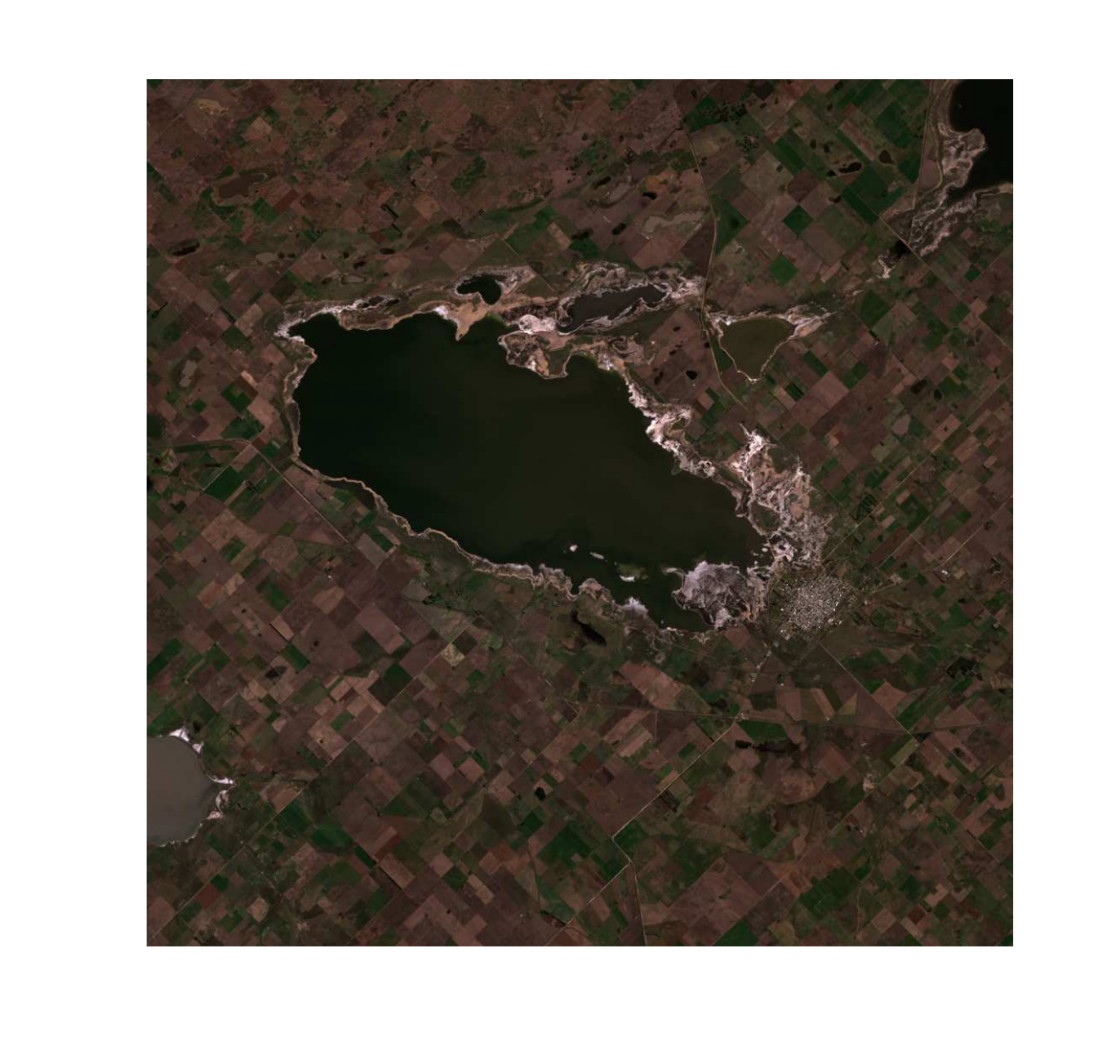}\label{colorLC8227086}}
        \subfigure[Classification with multifractal analysis.]{\includegraphics[width=0.4\textwidth]{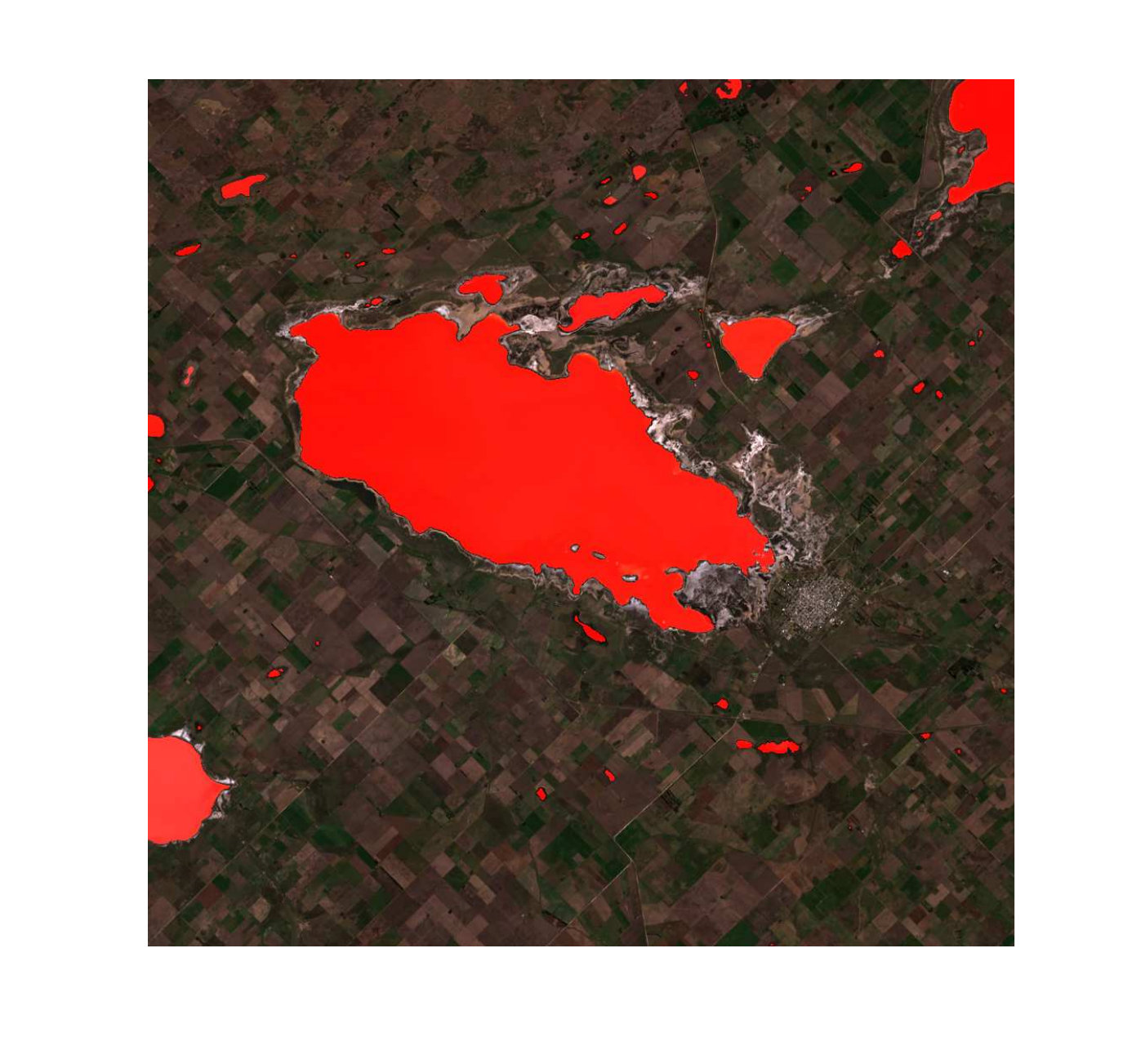}\label{multifractalLC8227086}}\\
        \subfigure[Classification with NDWI.]{\includegraphics[width=0.4\textwidth]{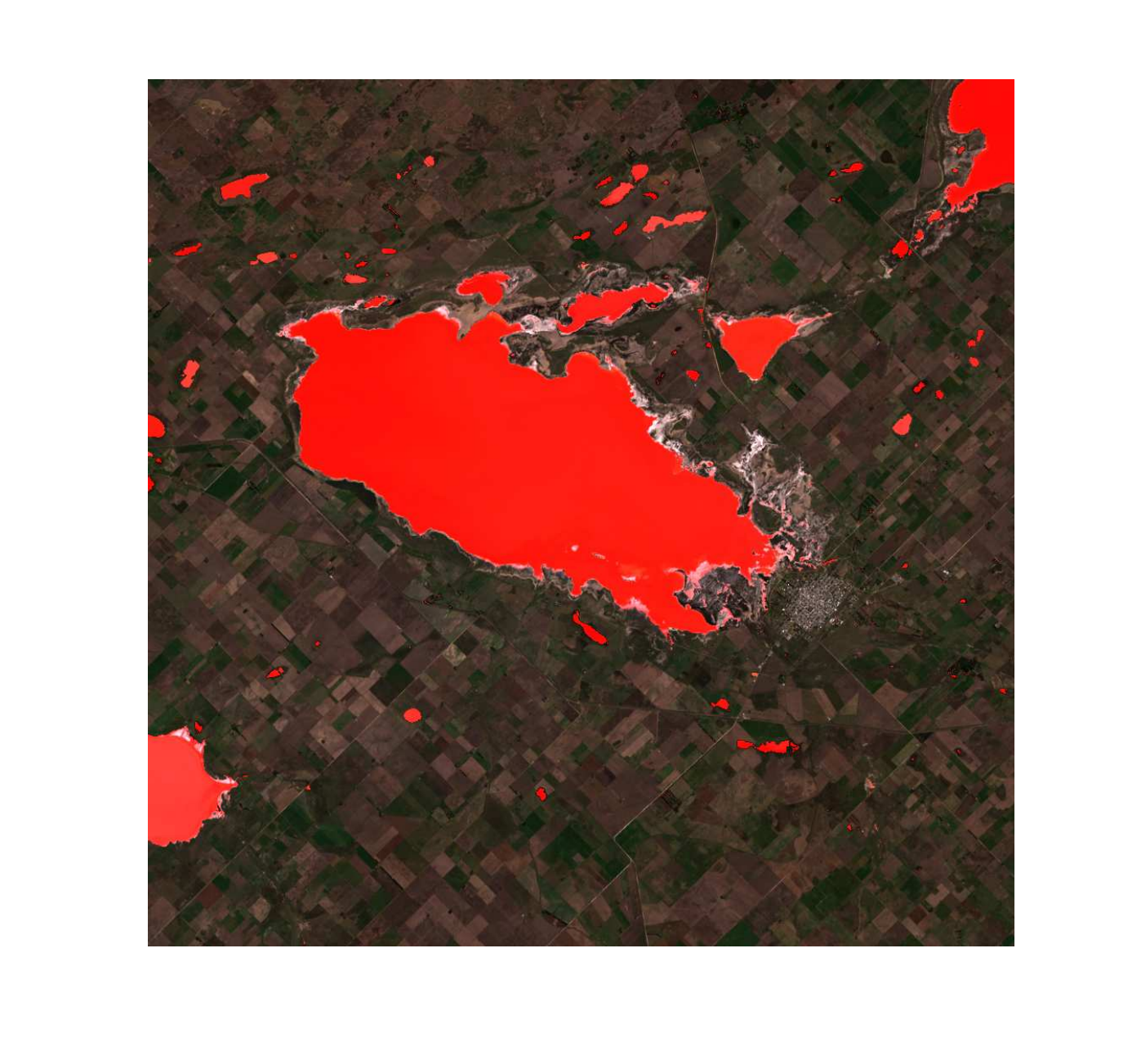}\label{ndwiLC8227086}}
        \subfigure[Classification with neural networks.]{\includegraphics[width=0.4\textwidth]{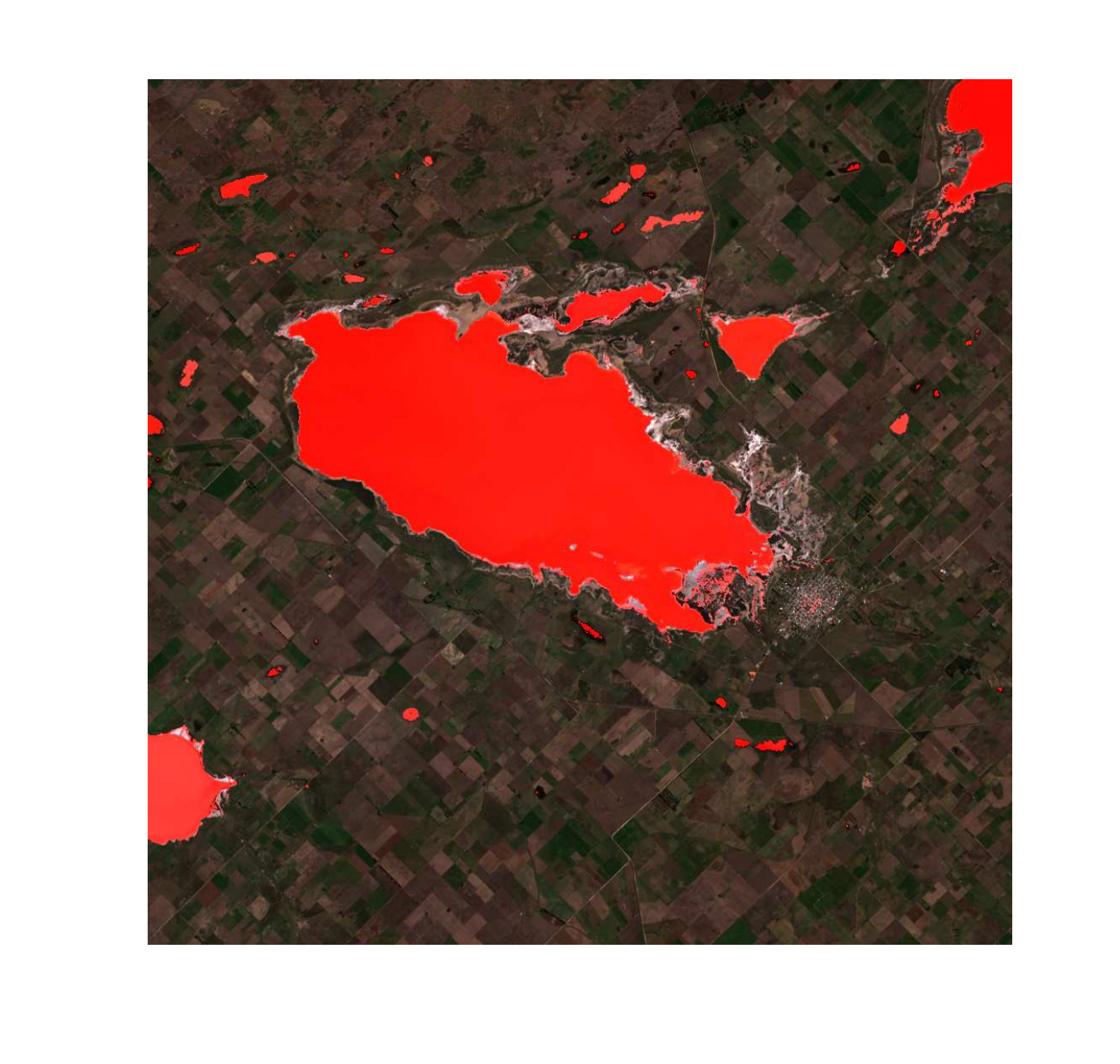}\label{neuronaLC8227086}}
        \caption{Comparison of the results of classification of the different methods on the image LC8227086.}\label{clasificacionLC8227086}
\end{figure}

\begin{figure}
\centering
        \subfigure[True color image.]{\includegraphics[width=0.4\textwidth]{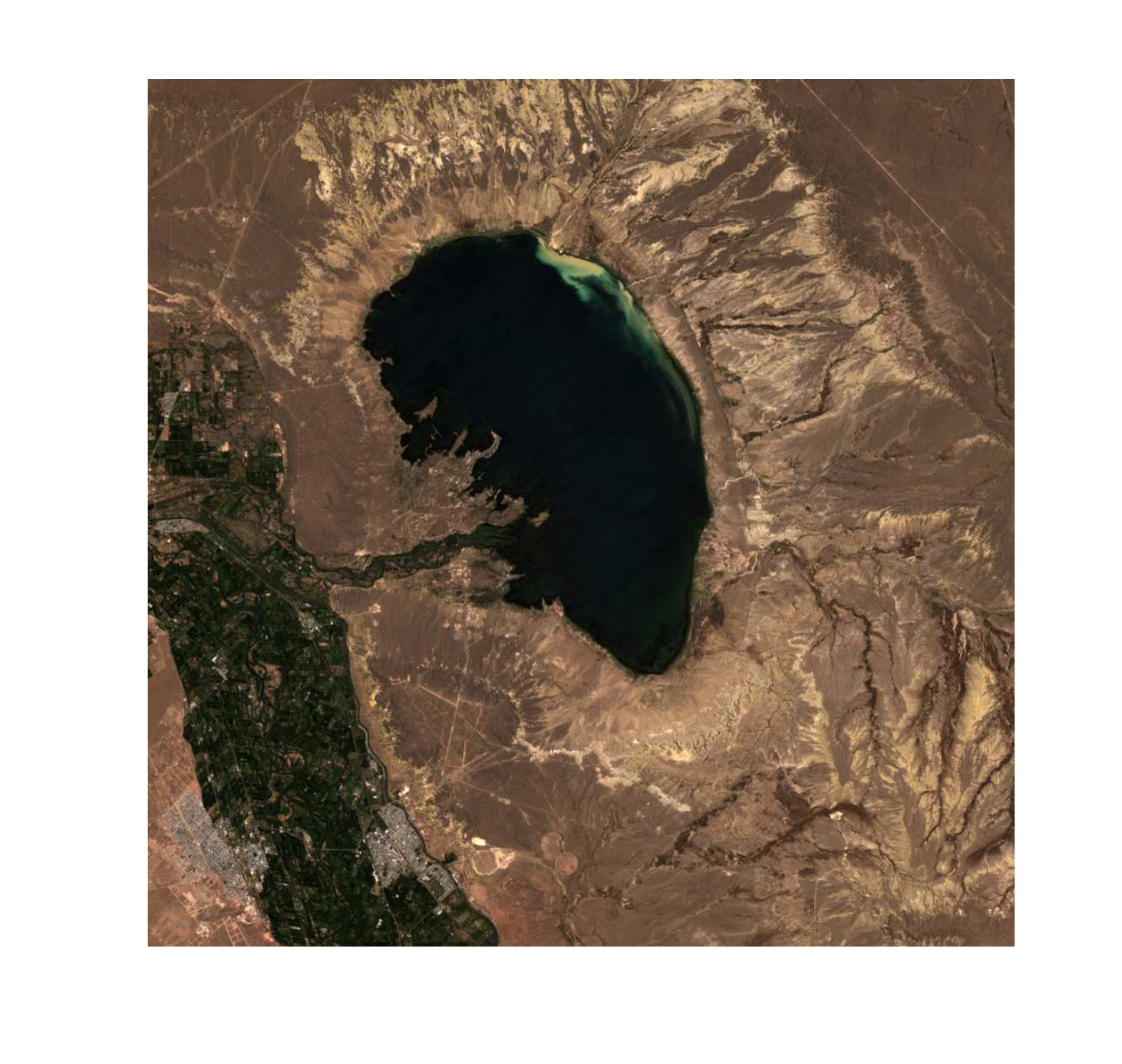}\label{colorLC8230087}}
        \subfigure[Classification with multifractal analysis.]{\includegraphics[width=0.4\textwidth]{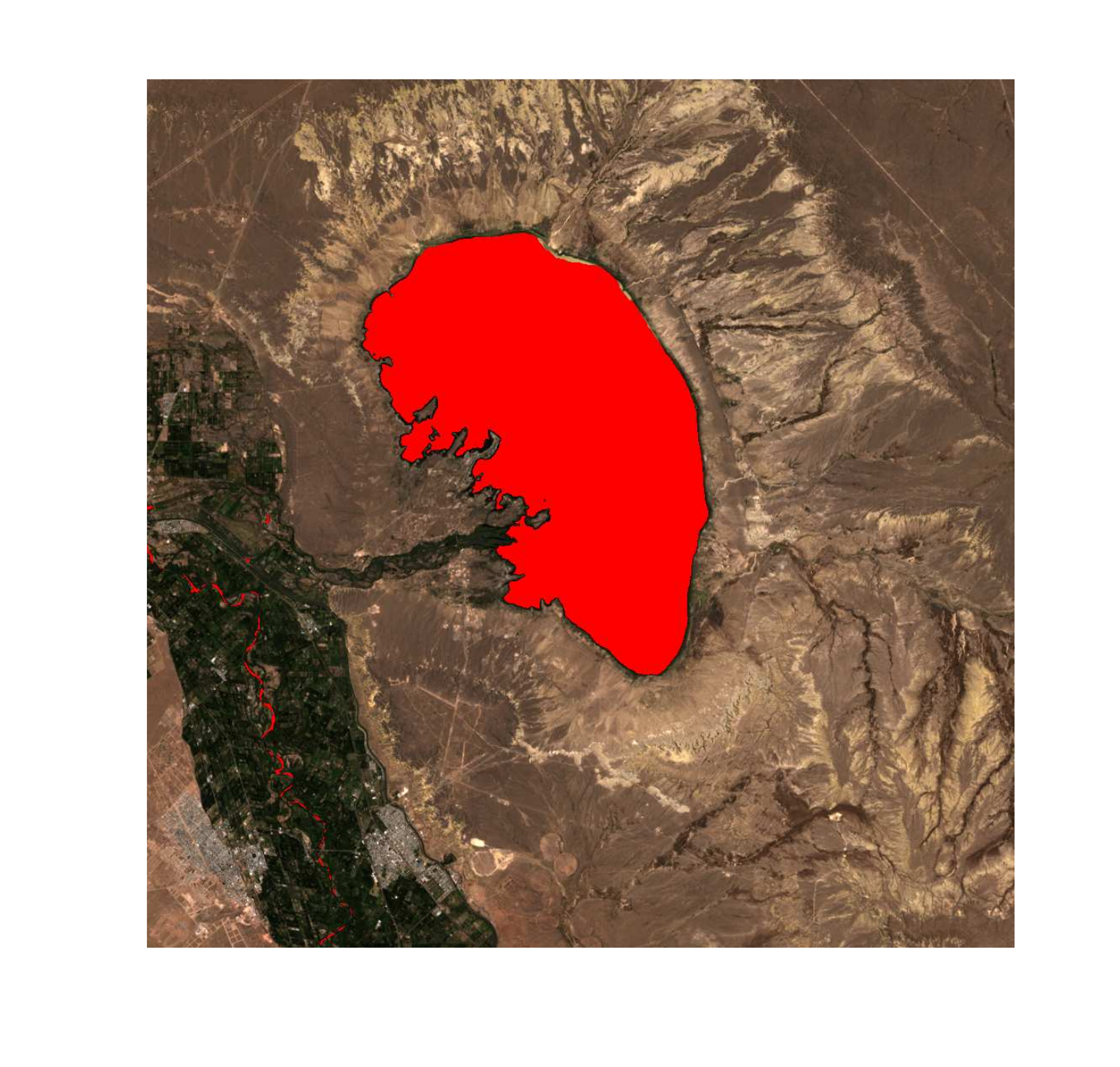}\label{multifractalLC8230087}}\\
        \subfigure[Classification with NDWI.]{\includegraphics[width=0.4\textwidth]{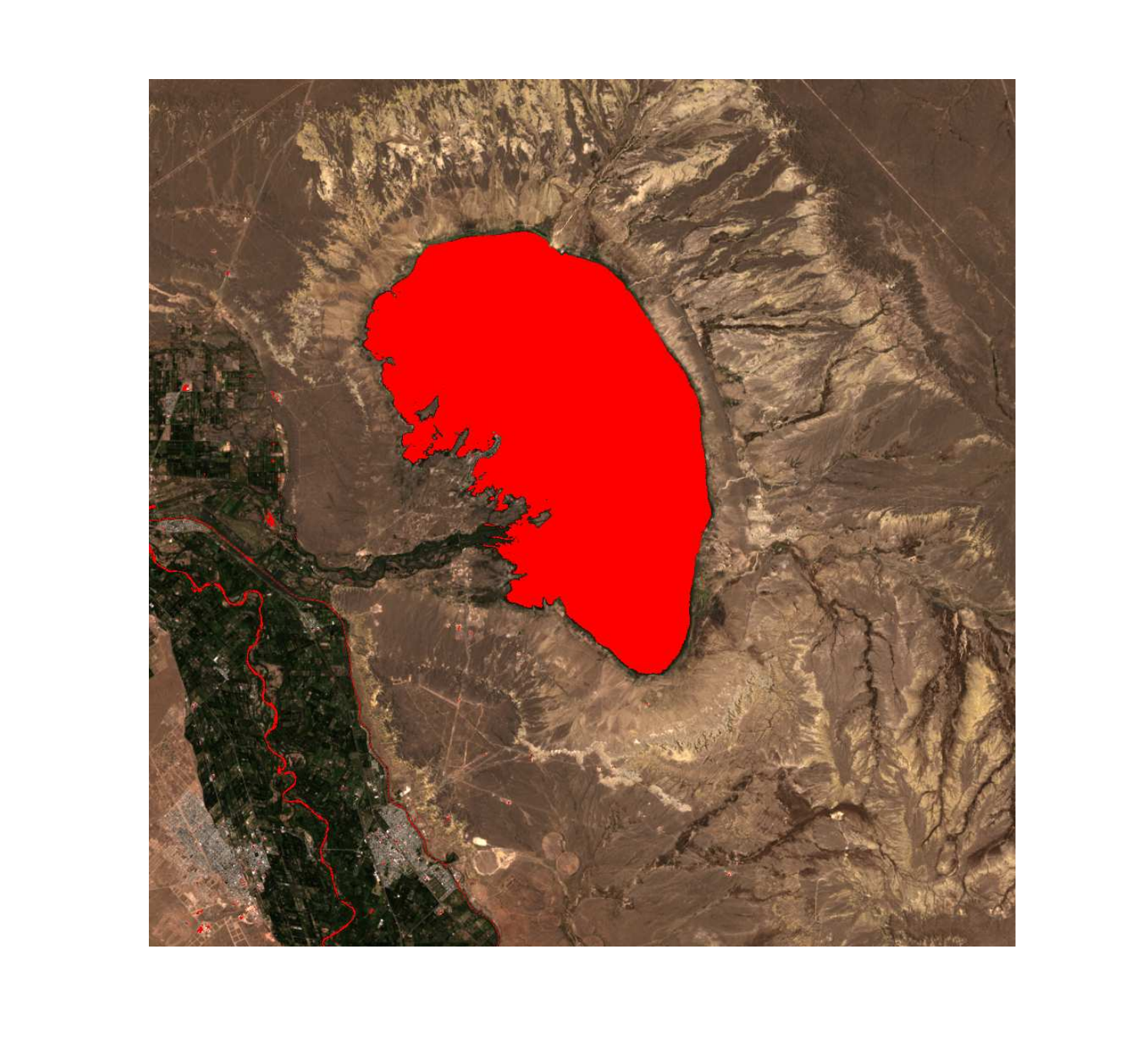}\label{ndwiLC8230087}}
        \subfigure[Classification with neural networks.]{\includegraphics[width=0.4\textwidth]{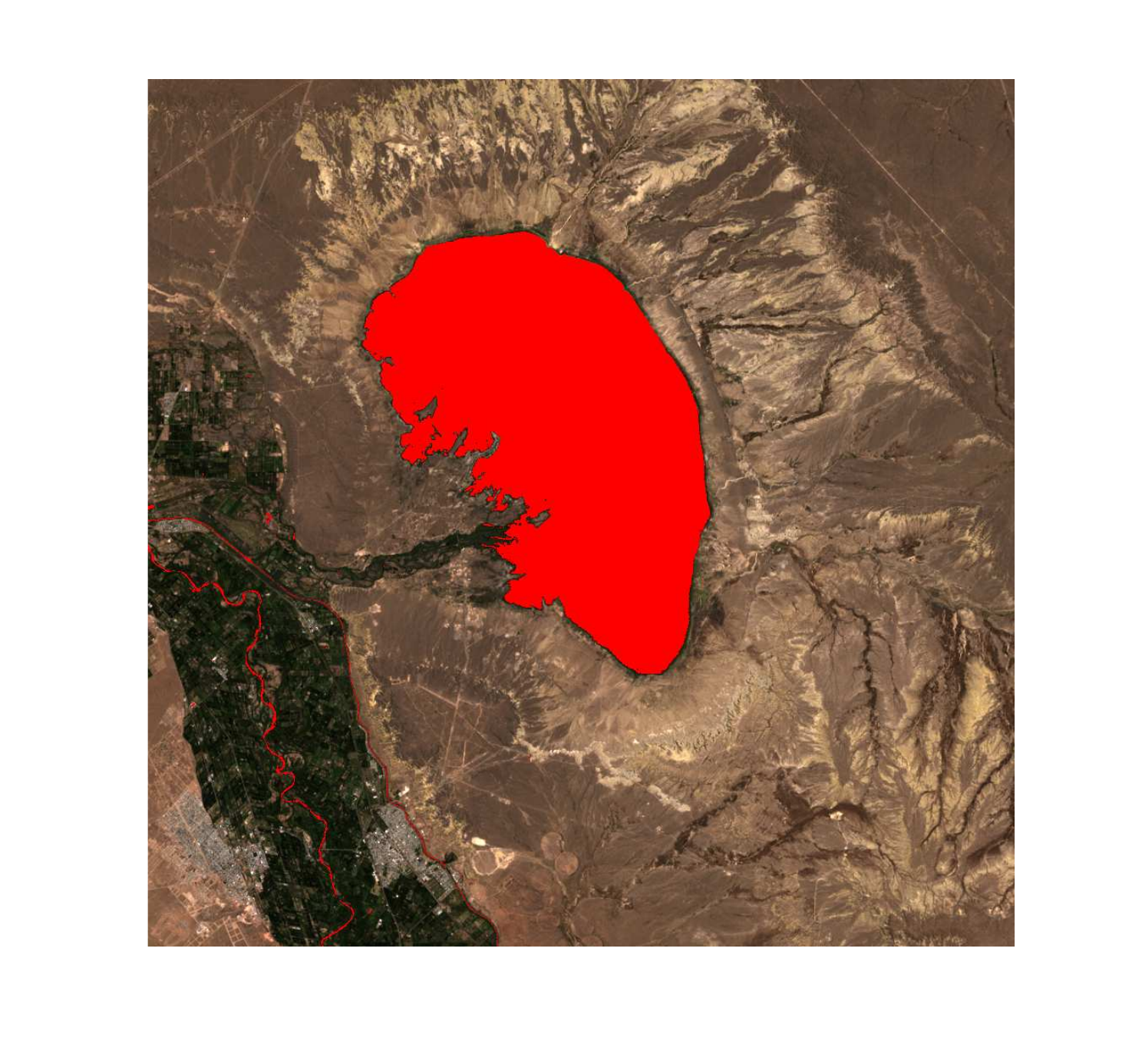}\label{neuronaLC8230087}}
        \caption{Comparison of the results of classification of the different methods on the image LC8230087.}\label{clasificacionLC8230087}
\end{figure}

Exploratory study of the graphs of multifractal spectra revealed that spectra obtained do not have a uniform concavity as those obtained statistically (figures \ref{fractalcomparacionLC8230087} and \ref{fractalcomparacionS1005959}), but show a depression in its central region. For values of $\alpha$ corresponding to the depression, $f(\alpha)\approx1$. That would indicate the presence of a one-dimensional structure separating two regions of different local regularity. These regions, with local maxima in which $f(\alpha)>1$ correspond to sets of pixels with a similar regularity.

This particular form of the spectrum, along with the hypothesis that the regions covered with water are more regular than those with different soil types, led us to select the area of the multifractal spectrum with values of $\alpha$ greater than those for the depression in the graph. On the other hand, the presence of different values of $f(\alpha)$ for the values of $\alpha$  in this region would indicate the existence of image areas with different structures, some of them with a dimension close to the one of a surface, others like lines and some others more localized. This variety may be due to the existence of various geographical features like lakes, rivers and streams of water, etc. We decided to include all of these possible structures within the classification. Therefore the selection of the thresholds includes values of $\alpha$ greater than the corresponding to the depression of the graph and values of $f(\alpha)$ smaller than the local maximum in this region of the spectrum.

It is important to highlight the rather complex concavity properties of all the  multifractal spectra obtained for these images using the coarse theory. We are not aware of similar examples in the literature and we suspect this could be a generic property of multifractal spectra of Earth images and perhaps even more general. The first immediate consequence is that the multifractal spectra derived from the partition function will lose most of the action.

This is interesting conceptually, because the method by which this spectrum is obtained not only uses global information of the image, but also has a local approach at different scales. Then, this approach gives us more information than an entirely statistical analysis, identifying details that would be lost otherwise.

From a practical point of view, the importance of this feature is the correlation between these subsets and distinct land covers. This situation allows us to apply thresholds to multifractal spectrum, thus performing an efficient image segmentation.

\begin{figure}
\centering
  \includegraphics[width=0.9\linewidth]{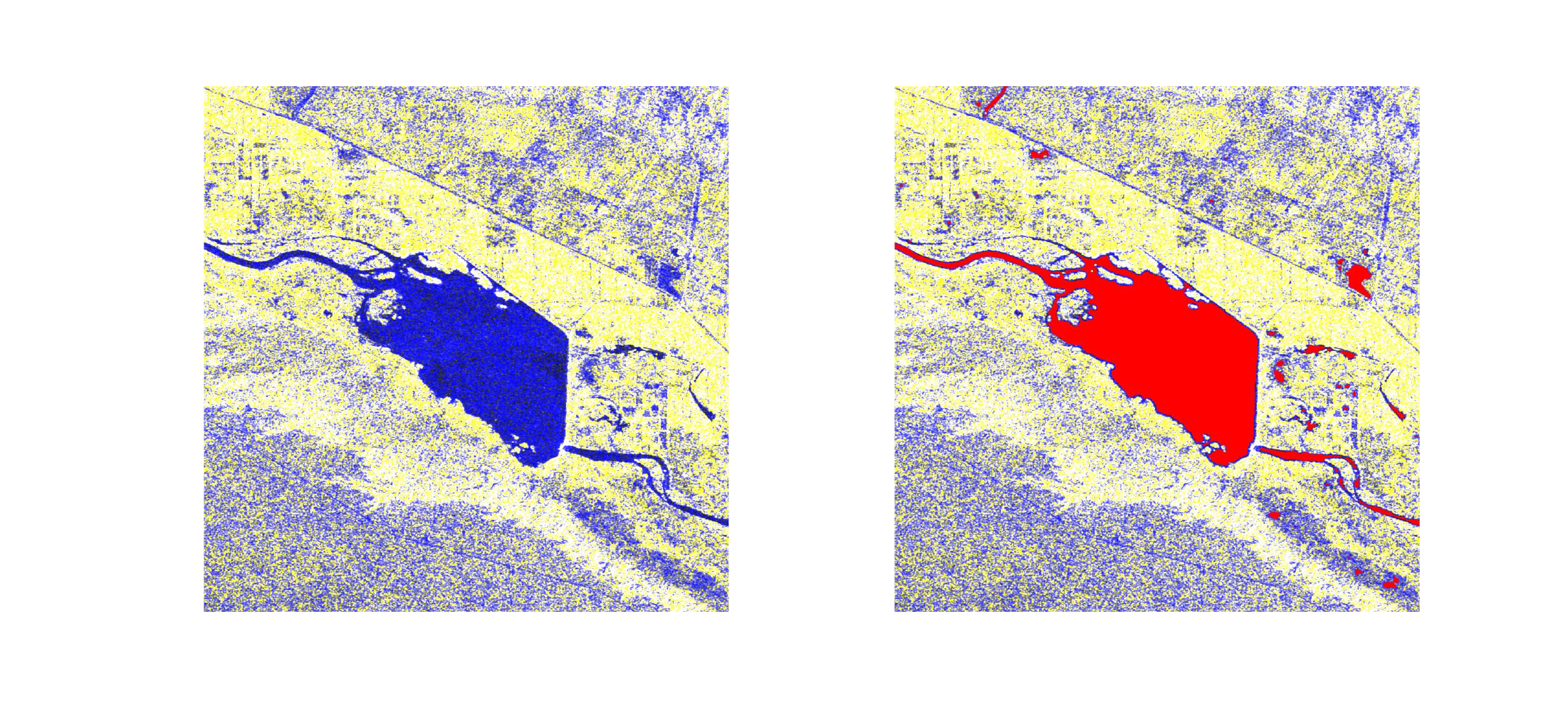}
  \caption{Comparison of the false-color composite image S1005959 with one in which the region detected as water by means of multifractal analysis is highlighted.}\label{fractalsegmentacioncomparadaS1005959}
  \includegraphics[width=0.9\linewidth]{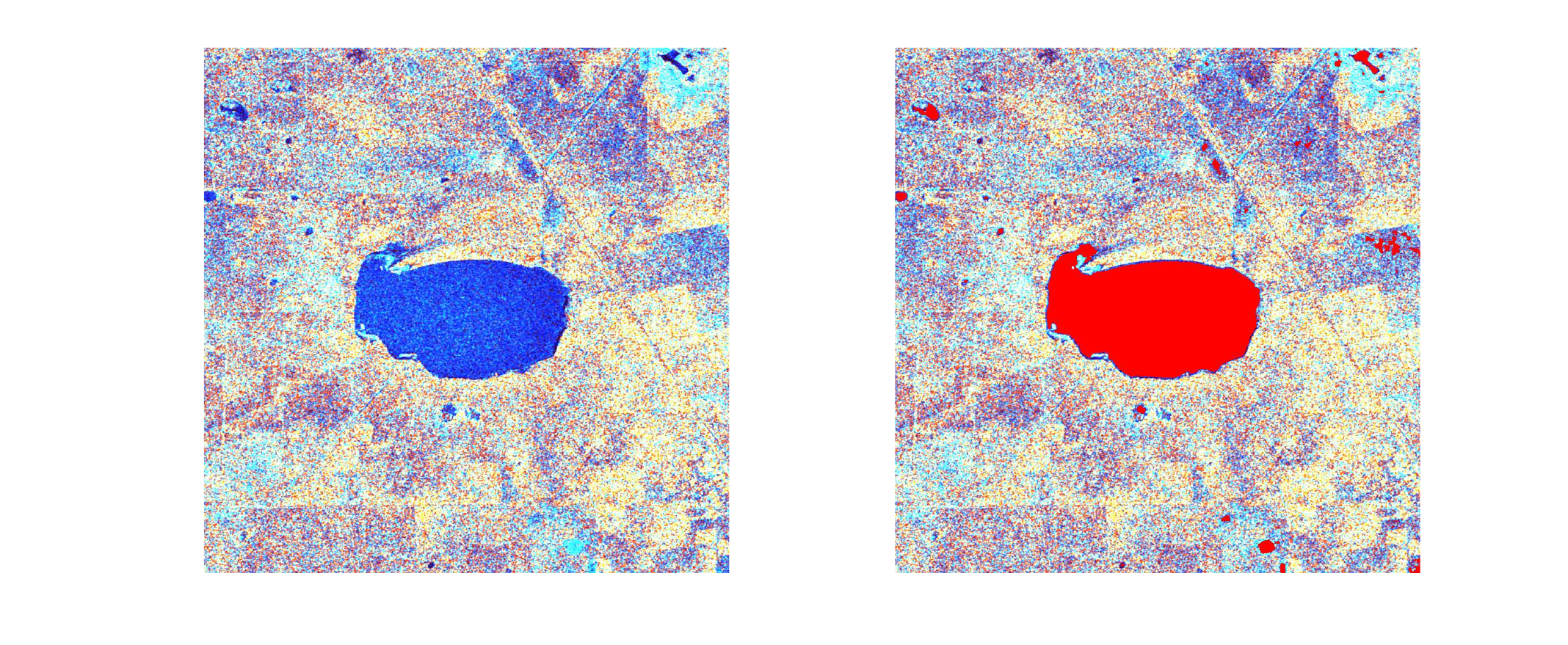}
  \caption{Comparison of the false-color composite image S1005886 with one in which the region detected as water due to the application of thresholds $0<f(\alpha)<1.39$ and $2.21<\alpha<2.54$ is highlighted.}\label{fractalsegmentacioncomparadaS1005886}
\end{figure}

The multifractal spectra presented in figure \ref{fractalcomparacionLC8230087} are obtained as a result of analysis on the figure LC8230087. The range of H\"{o}lder exponents $[1.75; 2.55]$ was obtained from  coarse multifractal spectrum. The thresholds used on the spectrum for classification are $0<f(\alpha)<1.38$ and $2.15<\alpha<\alpha_{max}=2.55$.

Values of $\alpha$ in $[1.75; 2.56]$ were observed for image LC8227086. The thresholds $0<f(\alpha)<1.38$ and $2.15<\alpha<\alpha_{max}=2.56$ were used in classification.

The results of the classifications on LC8230087 and LC8227086 are compared with the original images in figures \ref{clasificacionLC8227086} and \ref{clasificacionLC8230087} respectively.

In order to quantitatively compare multifractal method classifications with the results of other classifications, confusion matrices were used. The confusion matrices give us indicators about the coincidence between the tested method (multifractal) and the reference method. They are:

{\footnotesize
\begin{eqnarray}
\mbox{Positive predictive value (PPV)}&=&\frac{\mbox{True positives}}{\mbox{True positives}+\mbox{False positives}}\times 100\\
\mbox{Negative predictive value (NPV)}&=&\frac{\mbox{True negatives}}{\mbox{True negatives}+\mbox{False negatives}}\times 100\\
\mbox{Sensitivity}&=&\frac{\mbox{True positives}}{\mbox{True positives}+\mbox{False negatives}}\times 100\\
\mbox{Specificity}&=&\frac{\mbox{True negatives}}{\mbox{True negatives}+\mbox{False positives}}\times 100\\
\mbox{Accuracy}&=&\frac{\mbox{True positives}+\mbox{True negatives}}{\mbox{Total population}}\times 100
\end{eqnarray}}

\begin{table}
\caption{Confusion matrices resulting from the comparison between the multifractal classification and classifications due to the neural network and NDWI for the image LC8227086.}\label{confusionLC8227086}

\begin{minipage}[c]{.5\linewidth}
    \begin{center}
       \scalebox{0.6}{{\setlength{\arrayrulewidth}{0.5mm}
       \begin{tabular}{c p{0.0625\textwidth}|p{0.05\textwidth}|p{0.05\textwidth}|p{0.1\textwidth}}
         \cline{3-4}
           &   & \multicolumn{2}{c@{\vrule height 12pt depth 4pt width 0pt}|}{Neural network} &   \\
         \hhline{~---~}
           & \multicolumn{1}{|c|}{{\cellcolor{brown!50}Total}} {\vrule height 0.6cm depth 0cm width 0pt}&  &  &   \\
           & \multicolumn{1}{|c|}{{\cellcolor{brown!50}population:}}{\vrule height 0cm depth 0cm width 0pt}& \multicolumn{1}{c|}{Water}& \multicolumn{1}{c|}{Not water}&   \\
           & \multicolumn{1}{|c|}{{\cellcolor{brown!50}1048576}}{\vrule height 0cm depth 0.6cm width 0pt}& & &   \\
         \hline
         \multicolumn{1}{|c}{\multirow{4}{*}{\begin{sideways}\mbox{Multifractal analysis}\end{sideways}}} & \multicolumn{1}{|c|}{ Water }{\vrule height 1cm depth 0cm width 0pt}& \multicolumn{1}{c|}{ \cellcolor{green!50}138998} & \multicolumn{1}{c|}{\cellcolor{red!50}1901} & \multicolumn{1}{c|}{ {\cellcolor{gray!50}PPV:}} \\
         \multicolumn{1}{|c}{ }{\vrule height 0cm depth 0.8cm width 0pt}&\multicolumn{1}{|c|}{ }& \multicolumn{1}{c|}{\cellcolor{green!50}(13.26\%)} & \multicolumn{1}{c|}{\cellcolor{red!50}(0.18\%)} & \multicolumn{1}{c|}{{\cellcolor{gray!50}98.65\%}} \\
         \hhline{~----}
        \multicolumn{1}{|c}{ }{\vrule height 1cm depth 0cm width 0pt}& \multicolumn{1}{|c|}{ Not water }  & \multicolumn{1}{c|}{{\cellcolor{red!50}14972}} & \multicolumn{1}{c|}{{\cellcolor{green!50}892705}}  & \multicolumn{1}{c|}{{\cellcolor{gray!50}NPV:}} \\
        \multicolumn{1}{|c}{ }{\vrule height 0cm depth 0.8cm width 0pt}&\multicolumn{1}{|c|}{ }& \multicolumn{1}{c|}{{\cellcolor{red!50}(1.43\%)}}  & \multicolumn{1}{c|}{{\cellcolor{green!50}(85.13\%)}} & \multicolumn{1}{c|}{{\cellcolor{gray!50}98.35\%}} \\
         \hline
           \multicolumn{1}{c}{ }{\vrule height 1cm depth 0cm width 0pt}& \multicolumn{1}{c|}{} & \multicolumn{1}{c|}{{\cellcolor{gray!50}Sensitivity:}}& \multicolumn{1}{c|}{{\cellcolor{gray!50}Specificity:}}& \multicolumn{1}{c|}{{\cellcolor{blue!50}Accuracy:}} \\
           \multicolumn{1}{c}{ }{\vrule height 0cm depth 0.8cm width 0pt}&\multicolumn{1}{c|}{ }&  \multicolumn{1}{c|}{{\cellcolor{gray!50}90.28\%}} & \multicolumn{1}{c|}{{\cellcolor{gray!50}99.79\%}} & \multicolumn{1}{c|}{{\cellcolor{blue!50}98.39\%}} \\
         \cline{3-5}
       \end{tabular}}}
    \end{center}
\end{minipage}
\hfill
\begin{minipage}[c]{.5\linewidth}
    \begin{center}
       \scalebox{0.6}{{\setlength{\arrayrulewidth}{0.5mm}\begin{tabular}{c p{0.0625\textwidth}|p{0.05\textwidth}|p{0.05\textwidth}|p{0.1\textwidth}}
         \cline{3-4}
           &   & \multicolumn{2}{c@{\vrule height 12pt depth 4pt width 0pt}|}{NDWI} & \\
         \hhline{~---~}
           & \multicolumn{1}{|c|}{{\cellcolor{brown!50}Total}} {\vrule height 0.6cm depth 0cm width 0pt}&  &  &   \\
           & \multicolumn{1}{|c|}{{\cellcolor{brown!50}population:}}{\vrule height 0cm depth 0cm width 0pt}& \multicolumn{1}{c|}{Water}& \multicolumn{1}{c|}{Not water}&   \\
           & \multicolumn{1}{|c|}{{\cellcolor{brown!50}1048576}}{\vrule height 0cm depth 0.6cm width 0pt}& & &   \\
         \hline
         \multicolumn{1}{|c}{\multirow{4}{*}{\begin{sideways}\mbox{Multifractal analysis}\end{sideways}}} & \multicolumn{1}{|c|}{ Water }{\vrule height 1cm depth 0cm width 0pt}& \multicolumn{1}{c|}{ \cellcolor{green!50}139584} & \multicolumn{1}{c|}{\cellcolor{red!50}368} & \multicolumn{1}{c|}{ {\cellcolor{gray!50}PPV:}} \\
         \multicolumn{1}{|c}{ }{\vrule height 0cm depth 0.8cm width 0pt}&\multicolumn{1}{|c|}{ }& \multicolumn{1}{c|}{\cellcolor{green!50}(13.31\%)} & \multicolumn{1}{c|}{\cellcolor{red!50}(0.04\%)} & \multicolumn{1}{c|}{{\cellcolor{gray!50}99.74\%}} \\
         \hhline{~----}
        \multicolumn{1}{|c}{ }{\vrule height 1cm depth 0cm width 0pt}& \multicolumn{1}{|c|}{ Not water }  & \multicolumn{1}{c|}{{\cellcolor{red!50}17111}} & \multicolumn{1}{c|}{{\cellcolor{green!50}891513}}  & \multicolumn{1}{c|}{{\cellcolor{gray!50}NPV:}} \\
        \multicolumn{1}{|c}{ }{\vrule height 0cm depth 0.8cm width 0pt}&\multicolumn{1}{|c|}{ }& \multicolumn{1}{c|}{{\cellcolor{red!50}(1.63\%)}}  & \multicolumn{1}{c|}{{\cellcolor{green!50}(85.02\%)}} & \multicolumn{1}{c|}{{\cellcolor{gray!50}98.12\%}} \\
         \hline
           \multicolumn{1}{c}{ }{\vrule height 1cm depth 0cm width 0pt}& \multicolumn{1}{c|}{} & \multicolumn{1}{c|}{{\cellcolor{gray!50}Sensitivity:}}& \multicolumn{1}{c|}{{\cellcolor{gray!50}Specificity:}}& \multicolumn{1}{c|}{{\cellcolor{blue!50}Accuracy:}} \\
           \multicolumn{1}{c}{ }{\vrule height 0cm depth 0.8cm width 0pt}&\multicolumn{1}{c|}{ }&  \multicolumn{1}{c|}{{\cellcolor{gray!50}89.08\%}} & \multicolumn{1}{c|}{{\cellcolor{gray!50}99.96\%}} & \multicolumn{1}{c|}{{\cellcolor{blue!50}98.33\%}} \\
         \cline{3-5}
       \end{tabular}}}
    \end{center}
\end{minipage}
\end{table}

\begin{table}
\caption{Confusion matrices resulting from the comparison between the multifractal classification and classifications due to the neural network and NDWI for the image LC8230087.}\label{confusionLC8230087}
\begin{minipage}[c]{.5\linewidth}
    \begin{center}
       \scalebox{0.6}{{\setlength{\arrayrulewidth}{0.5mm}\begin{tabular}{c p{0.0625\textwidth}|p{0.05\textwidth}|p{0.05\textwidth}|p{0.1\textwidth}}
         \cline{3-4}
           &   & \multicolumn{2}{c@{\vrule height 12pt depth 4pt width 0pt}|}{Neural network} &   \\
         \hhline{~---~}
           & \multicolumn{1}{|c|}{{\cellcolor{brown!50}Total}} {\vrule height 0.6cm depth 0cm width 0pt}&  &  &   \\
           & \multicolumn{1}{|c|}{{\cellcolor{brown!50}population:}}{\vrule height 0cm depth 0cm width 0pt}& \multicolumn{1}{c|}{Water}& \multicolumn{1}{c|}{Not water}&   \\
           & \multicolumn{1}{|c|}{{\cellcolor{brown!50}1048576}}{\vrule height 0cm depth 0.6cm width 0pt}& & &   \\
         \hline
         \multicolumn{1}{|c}{\multirow{4}{*}{\begin{sideways}\mbox{Multifractal analysis}\end{sideways}}} & \multicolumn{1}{|c|}{ Water }{\vrule height 1cm depth 0cm width 0pt}& \multicolumn{1}{c|}{ \cellcolor{green!50}125208} & \multicolumn{1}{c|}{\cellcolor{red!50}93} & \multicolumn{1}{c|}{ {\cellcolor{gray!50}PPV:}} \\
         \multicolumn{1}{|c}{ }{\vrule height 0cm depth 0.8cm width 0pt}&\multicolumn{1}{|c|}{ }& \multicolumn{1}{c|}{\cellcolor{green!50}(11.94\%)} & \multicolumn{1}{c|}{\cellcolor{red!50}(0.01\%)} & \multicolumn{1}{c|}{{\cellcolor{gray!50}99.93\%}} \\
         \hhline{~----}
        \multicolumn{1}{|c}{ }{\vrule height 1cm depth 0cm width 0pt}& \multicolumn{1}{|c|}{ Not water }  & \multicolumn{1}{c|}{{\cellcolor{red!50}4601}} & \multicolumn{1}{c|}{{\cellcolor{green!50}918674}}  & \multicolumn{1}{c|}{{\cellcolor{gray!50}NPV:}} \\
        \multicolumn{1}{|c}{ }{\vrule height 0cm depth 0.8cm width 0pt}&\multicolumn{1}{|c|}{ }& \multicolumn{1}{c|}{{\cellcolor{red!50}(0.44\%)}}  & \multicolumn{1}{c|}{{\cellcolor{green!50}(87.61\%)}} & \multicolumn{1}{c|}{{\cellcolor{gray!50}99.50\%}} \\
         \hline
           \multicolumn{1}{c}{ }{\vrule height 1cm depth 0cm width 0pt}& \multicolumn{1}{c|}{} & \multicolumn{1}{c|}{{\cellcolor{gray!50}Sensitivity:}}& \multicolumn{1}{c|}{{\cellcolor{gray!50}Specificity:}}& \multicolumn{1}{c|}{{\cellcolor{blue!50}Accuracy:}} \\
           \multicolumn{1}{c}{ }{\vrule height 0cm depth 0.8cm width 0pt}&\multicolumn{1}{c|}{ }&  \multicolumn{1}{c|}{{\cellcolor{gray!50}96.46\%}} & \multicolumn{1}{c|}{{\cellcolor{gray!50}99.99\%}} & \multicolumn{1}{c|}{{\cellcolor{blue!50}99.55\%}} \\
         \cline{3-5}
       \end{tabular}}}
    \end{center}
\end{minipage}
\hfill
\begin{minipage}[c]{.5\linewidth}
    \begin{center}
       \scalebox{0.6}{{\setlength{\arrayrulewidth}{0.5mm}\begin{tabular}{c p{0.0625\textwidth}|p{0.05\textwidth}|p{0.05\textwidth}|p{0.1\textwidth}}
         \cline{3-4}
           &   & \multicolumn{2}{c@{\vrule height 12pt depth 4pt width 0pt}|}{NDWI} &   \\
         \hhline{~---~}
           & \multicolumn{1}{|c|}{{\cellcolor{brown!50}Total}} {\vrule height 0.6cm depth 0cm width 0pt}&  &  &   \\
           & \multicolumn{1}{|c|}{{\cellcolor{brown!50}population:}}{\vrule height 0cm depth 0cm width 0pt}& \multicolumn{1}{c|}{Water}& \multicolumn{1}{c|}{Not water}&   \\
           & \multicolumn{1}{|c|}{{\cellcolor{brown!50}1048576}}{\vrule height 0cm depth 0.6cm width 0pt}& & &   \\
         \hline
         \multicolumn{1}{|c}{\multirow{4}{*}{\begin{sideways}\mbox{Multifractal analysis}\end{sideways}}} & \multicolumn{1}{|c|}{ Water }{\vrule height 1cm depth 0cm width 0pt}& \multicolumn{1}{c|}{ \cellcolor{green!50}125202} & \multicolumn{1}{c|}{\cellcolor{red!50}99} & \multicolumn{1}{c|}{ {\cellcolor{gray!50}PPV:}} \\
         \multicolumn{1}{|c}{ }{\vrule height 0cm depth 0.8cm width 0pt}&\multicolumn{1}{|c|}{ }& \multicolumn{1}{c|}{\cellcolor{green!50}(11.94\%)} & \multicolumn{1}{c|}{\cellcolor{red!50}(0.01\%)} & \multicolumn{1}{c|}{{\cellcolor{gray!50}99.92\%}} \\
         \hhline{~----}
        \multicolumn{1}{|c}{ }{\vrule height 1cm depth 0cm width 0pt}& \multicolumn{1}{|c|}{ Not water }  & \multicolumn{1}{c|}{{\cellcolor{red!50}5257}} & \multicolumn{1}{c|}{{\cellcolor{green!50}918018}}  & \multicolumn{1}{c|}{{\cellcolor{gray!50}NPV:}} \\
        \multicolumn{1}{|c}{ }{\vrule height 0cm depth 0.8cm width 0pt}&\multicolumn{1}{|c|}{ }& \multicolumn{1}{c|}{{\cellcolor{red!50}(0.50\%)}}  & \multicolumn{1}{c|}{{\cellcolor{green!50}(87.55\%)}} & \multicolumn{1}{c|}{{\cellcolor{gray!50}99.43\%}} \\
         \hline
           \multicolumn{1}{c}{ }{\vrule height 1cm depth 0cm width 0pt}& \multicolumn{1}{c|}{} & \multicolumn{1}{c|}{{\cellcolor{gray!50}Sensitivity:}}& \multicolumn{1}{c|}{{\cellcolor{gray!50}Specificity:}}& \multicolumn{1}{c|}{{\cellcolor{blue!50}Accuracy:}} \\
           \multicolumn{1}{c}{ }{\vrule height 0cm depth 0.8cm width 0pt}&\multicolumn{1}{c|}{ }&  \multicolumn{1}{c|}{{\cellcolor{gray!50}95.97\%}} & \multicolumn{1}{c|}{{\cellcolor{gray!50}99.99\%}} & \multicolumn{1}{c|}{{\cellcolor{blue!50}99.45\%}} \\
         \cline{3-5}
       \end{tabular}}}
    \end{center}
\end{minipage}
\end{table}

\begin{table}
\begin{minipage}[c]{.49\linewidth}
    \centering
    \caption{Confusion matrices resulting from the comparison between the multifractal classification and classification due to the neural network for the image S1005886.}\label{S1005886}
    \begin{center}
       \scalebox{0.6}{{\setlength{\arrayrulewidth}{0.5mm}\begin{tabular}{c p{0.0625\textwidth}|p{0.05\textwidth}|p{0.05\textwidth}|p{0.1\textwidth}}
         \cline{3-4}
           &   & \multicolumn{2}{c@{\vrule height 12pt depth 4pt width 0pt}|}{Neural network} &   \\
         \hhline{~---~}
           & \multicolumn{1}{|c|}{{\cellcolor{brown!50}Total}} {\vrule height 0.6cm depth 0cm width 0pt}&  &  &   \\
           & \multicolumn{1}{|c|}{{\cellcolor{brown!50}population:}}{\vrule height 0cm depth 0cm width 0pt}& \multicolumn{1}{c|}{Water}& \multicolumn{1}{c|}{Not water}&   \\
           & \multicolumn{1}{|c|}{{\cellcolor{brown!50}1048576}}{\vrule height 0cm depth 0.6cm width 0pt}& & &   \\
         \hline
         \multicolumn{1}{|c}{\multirow{4}{*}{\begin{sideways}\mbox{Multifractal analysis}\end{sideways}}} & \multicolumn{1}{|c|}{ Water }{\vrule height 1cm depth 0cm width 0pt}& \multicolumn{1}{c|}{ \cellcolor{green!50}78500} & \multicolumn{1}{c|}{\cellcolor{red!50}4004} & \multicolumn{1}{c|}{ {\cellcolor{gray!50}PPV:}} \\
         \multicolumn{1}{|c}{ }{\vrule height 0cm depth 0.8cm width 0pt}&\multicolumn{1}{|c|}{ }& \multicolumn{1}{c|}{\cellcolor{green!50}(7.49\%)} & \multicolumn{1}{c|}{\cellcolor{red!50}(0.38\%)} & \multicolumn{1}{c|}{{\cellcolor{gray!50}95.15\%}} \\
         \hhline{~----}
        \multicolumn{1}{|c}{ }{\vrule height 1cm depth 0cm width 0pt}& \multicolumn{1}{|c|}{ Not water }  & \multicolumn{1}{c|}{{\cellcolor{red!50}3436}} & \multicolumn{1}{c|}{{\cellcolor{green!50}962636}}  & \multicolumn{1}{c|}{{\cellcolor{gray!50}NPV:}} \\
        \multicolumn{1}{|c}{ }{\vrule height 0cm depth 0.8cm width 0pt}&\multicolumn{1}{|c|}{ }& \multicolumn{1}{c|}{{\cellcolor{red!50}(0.33\%)}}  & \multicolumn{1}{c|}{{\cellcolor{green!50}(91.80\%)}} & \multicolumn{1}{c|}{{\cellcolor{gray!50}99.64\%}} \\
         \hline
           \multicolumn{1}{c}{ }{\vrule height 1cm depth 0cm width 0pt}& \multicolumn{1}{c|}{} & \multicolumn{1}{c|}{{\cellcolor{gray!50}Sensitivity:}}& \multicolumn{1}{c|}{{\cellcolor{gray!50}Specificity:}}& \multicolumn{1}{c|}{{\cellcolor{blue!50}Accuracy:}} \\
           \multicolumn{1}{c}{ }{\vrule height 0cm depth 0.8cm width 0pt}&\multicolumn{1}{c|}{ }&  \multicolumn{1}{c|}{{\cellcolor{gray!50}95.81\%}} & \multicolumn{1}{c|}{{\cellcolor{gray!50}99.59\%}} & \multicolumn{1}{c|}{{\cellcolor{blue!50}99.29\%}} \\
         \cline{3-5}
       \end{tabular}}}
    \end{center}
\end{minipage}
\hfill
\begin{minipage}[c]{.49\linewidth}
    \centering
    \caption{Confusion matrices resulting from the comparison between the multifractal classification and classification due to the neural network for the image S1005959.}\label{S1005959}
    \begin{center}
       \scalebox{0.6}{{\setlength{\arrayrulewidth}{0.5mm}\begin{tabular}{c p{0.0625\textwidth}|p{0.05\textwidth}|p{0.05\textwidth}|p{0.1\textwidth}}
         \cline{3-4}
           &   & \multicolumn{2}{c@{\vrule height 12pt depth 4pt width 0pt}|}{Neural network} &   \\
         \hhline{~---~}
           & \multicolumn{1}{|c|}{{\cellcolor{brown!50}Total}} {\vrule height 0.6cm depth 0cm width 0pt}&  &  &   \\
           & \multicolumn{1}{|c|}{{\cellcolor{brown!50}population:}}{\vrule height 0cm depth 0cm width 0pt}& \multicolumn{1}{c|}{Water}& \multicolumn{1}{c|}{Not water}&   \\
           & \multicolumn{1}{|c|}{{\cellcolor{brown!50}1048576}}{\vrule height 0cm depth 0.6cm width 0pt}& & &   \\
         \hline
         \multicolumn{1}{|c}{\multirow{4}{*}{\begin{sideways}\mbox{Multifractal analysis}\end{sideways}}} & \multicolumn{1}{|c|}{ Water }{\vrule height 1cm depth 0cm width 0pt}& \multicolumn{1}{c|}{ \cellcolor{green!50}93154} & \multicolumn{1}{c|}{\cellcolor{red!50}2735} & \multicolumn{1}{c|}{ {\cellcolor{gray!50}PPV:}} \\
         \multicolumn{1}{|c}{ }{\vrule height 0cm depth 0.8cm width 0pt}&\multicolumn{1}{|c|}{ }& \multicolumn{1}{c|}{\cellcolor{green!50}(8.88\%)} & \multicolumn{1}{c|}{\cellcolor{red!50}(0.26\%)} & \multicolumn{1}{c|}{{\cellcolor{gray!50}97.15\%}} \\
         \hhline{~----}
        \multicolumn{1}{|c}{ }{\vrule height 1cm depth 0cm width 0pt}& \multicolumn{1}{|c|}{ Not water }  & \multicolumn{1}{c|}{{\cellcolor{red!50}8740}} & \multicolumn{1}{c|}{{\cellcolor{green!50}943947}}  & \multicolumn{1}{c|}{{\cellcolor{gray!50}NPV:}} \\
        \multicolumn{1}{|c}{ }{\vrule height 0cm depth 0.8cm width 0pt}&\multicolumn{1}{|c|}{ }& \multicolumn{1}{c|}{{\cellcolor{red!50}(0.83\%)}}  & \multicolumn{1}{c|}{{\cellcolor{green!50}(90.02\%)}} & \multicolumn{1}{c|}{{\cellcolor{gray!50}99.08\%}} \\
         \hline
           \multicolumn{1}{c}{ }{\vrule height 1cm depth 0cm width 0pt}& \multicolumn{1}{c|}{} & \multicolumn{1}{c|}{{\cellcolor{gray!50}Sensitivity:}}& \multicolumn{1}{c|}{{\cellcolor{gray!50}Specificity:}}& \multicolumn{1}{c|}{{\cellcolor{blue!50}Accuracy:}} \\
           \multicolumn{1}{c}{ }{\vrule height 0cm depth 0.8cm width 0pt}&\multicolumn{1}{c|}{ }&  \multicolumn{1}{c|}{{\cellcolor{gray!50}94.42\%}} & \multicolumn{1}{c|}{{\cellcolor{gray!50}99.71\%}} & \multicolumn{1}{c|}{{\cellcolor{blue!50}98.91\%}} \\
         \cline{3-5}
       \end{tabular}}}
    \end{center}
\end{minipage}
\end{table}

Confusion matrices resulting from comparison between multifractal classification and the ones due to reference methods for images LC8227086 and LC8230087 are presented in tables \ref{confusionLC8227086} and \ref{confusionLC8230087}.

Multifractal spectra obtained for image S1005959 are shown in figure \ref{fractalcomparacionS1005959}. The range of values of $\alpha$ obtained from coarse multifractal spectrum was $[1.49; 2.68]$. In figure \ref{fractalsegmentacioncomparadaS1005959} one can see the classification resulting from the application of the thresholds $0<f(\alpha)<1.39$ and $2.3<\alpha<\alpha_{max}=2.68$.

For image S1005886, values of $\alpha$ into the set $[1.22; 2.54]$ were found. Pixels for that $0<f(\alpha)<1.39$  and $2.21<\alpha<\alpha_{max}=2.54$ were classified as water. The comparison between the classified and the original image is shown in figure \ref{fractalsegmentacioncomparadaS1005886}.

In figure \ref{neuronasegmentacionpuraS1005886} we can see the classification resulting from the neural network applied to the image, while the figure \ref{neuronasegmentacionfiltradaS1005886} shows the result of filtering it.

Tables \ref{S1005886} and \ref{S1005959} present confusion matrices associated with the comparison between the classifications for S1005886 and S1005959 images.

\section{Conclusions}\label{conclusiones}

Coarse multifractal spectrum turns out to be a useful tool to study water bodies segmentation in satellite images.

For optical images, the classification method based on the multifractal spectrum presented high coincidence with the results of neural networks and NDWI. In the confusion matrices one can see that the Sensitivity, Specificity, Positive predictive value, Negative predictive value and Accuracy have values over $89 \%$.

Neural networks did not provide a proper classification of radar images, probably because of high sensitivity to speckle noise. Because of that, the result had to be filtered, obtaining a product of better quality. The comparison between that product and the classification performed through multifractal analysis showed again a good coincidence. Confusion matrices indicators have values over $90 \%$.

These results allow us to conclude that coarse multifractal analysis is a useful tool to study water coverage on images. Two observations are in order: 1) We had to use both the local H\"{o}lder exponent $\alpha$ and the global information contained in the spectrum $f(\alpha)$ to derive useful thresholds for segmentation. Therefore a pure statistical approach like the partition method would have not been enough. 2) It is important to highlight the similar shape and at the same time the rather complex concavity properties of all the multifractal spectra derived from these images using the coarse theory. In particular, the depression in the midrange. We are not aware of similar examples in the literature and we suspect this could be a generic property of multifractal spectra of Earth images and perhaps even more general. As a consequence, the multifractal spectra derived from the partition function is different from the one derived using the coarse theory and in particular misses the depression.

Our results lead us to expect that a multifractal analysis as the one here presented with, perhaps, a different measure $\mu$ could be sensitive to other different land covers. This could lead to a series of measures whose multifractal properties allow a complete segmentation of satellite images.

\section*{Acknowledgments}

The authors would like to thank Miguel A. Virasoro for his invaluable contribution during the design of the study and in the analysis of the results. The authors thank CONICET and UNGS.

\section*{References}

\bibliographystyle{elsarticle-num}
\bibliography{biblio}

\end{document}